%% file: arxiv.tex
\documentclass[10pt,journal,compsoc]{IEEEtran}

\ifCLASSOPTIONcompsoc
  \usepackage[nocompress]{cite}
\else
  \usepackage{cite}
\fi

\ifCLASSINFOpdf
\else
\fi

\usepackage{caption}
\usepackage{url}
\usepackage{lmodern}
\usepackage{times}
\usepackage{epsfig}
\usepackage{graphicx}
\usepackage{amsmath}
\usepackage{amssymb}
\usepackage{multirow}
\usepackage{colortbl}
\usepackage{color}
\usepackage{threeparttable}
\usepackage{booktabs}
\usepackage{adjustbox}
\usepackage{subfig}
\usepackage[misc]{ifsym}
\usepackage{xcolor,colortbl}
\usepackage{makecell}
\usepackage{graphicx}
\usepackage{tikz}
\usepackage{overpic}
\usepackage{textcomp}
\usetikzlibrary{arrows.meta, positioning, shapes.geometric, calc}
\usepackage{xcolor}
\usepackage{amsmath}
\usepackage{wasysym}
\newcommand{\pub}[1]{\color{gray}{\scriptsize{[{#1}]}}}

\newcommand*{\belowrulesepcolor}[1]{%
  \noalign{%
    \kern-\belowrulesep 
    \begingroup 
      \color{#1}%
      \hrule height\belowrulesep 
    \endgroup 
    \vspace{-0.03mm}
  }%
} 
\newcommand*{\aboverulesepcolor}[1]{%
  \noalign{%
  \vspace{-0.03mm}
    \begingroup 
      \color{#1}%
      \hrule height\aboverulesep 
    \endgroup 
    \kern-\aboverulesep 
  }%
}

\usepackage{pifont}

\usepackage{xspace}
\makeatletter
\DeclareRobustCommand\onedot{\futurelet\@let@token\@onedot}
\def\@onedot{\ifx\@let@token.\else.\null\fi\xspace}
\def\eg{\emph{e.g}\onedot} 
\def\ie{\emph{i.e}\onedot} 
 
\def\etc{\emph{etc}\onedot} \def\vs{\emph{vs}\onedot}
 
\def\etal{\emph{et al}\onedot}

\usepackage[linesnumbered,lined,boxed,commentsnumbered,ruled]{algorithm2e}
\usepackage[pagebackref=true,breaklinks=true,colorlinks,bookmarks=false]{hyperref}
\def\etal{\textit{et al.}}
\definecolor{mygray}{gray}{.90}
\definecolor{mygray1}{gray}{.7}
\definecolor{mygray2}{gray}{.93}

\usepackage[capitalize]{cleveref}
\crefname{section}{Sec.}{Secs.}
\Crefname{section}{Section}{Sections}
\Crefname{table}{TABLE}{TABLES}
\crefname{table}{TABLE}{TABLES}

\begin{document}

\title{Multimodal Referring Segmentation: A Survey}

\author{Henghui Ding, Song Tang, Shuting He, Chang Liu, Zuxuan Wu, Yu-Gang Jiang, \IEEEmembership{Fellow,~IEEE}
\IEEEcompsocitemizethanks{
\IEEEcompsocthanksitem H. Ding, S. Tang, Z. Wu, Y.G. Jiang are with Fudan University, China. henghui.ding@gmail.com 
\IEEEcompsocthanksitem S. He is with Shanghai University of Finance and Economics, China.
\IEEEcompsocthanksitem C. Liu is with ByteDance Inc.
}
}

\markboth{}
{Shell \MakeLowercase{\textit{et al.}}: Bare Advanced Demo of IEEEtran.cls for IEEE Computer Society Journals}

\input{Secs/0_abstract.tex}

\maketitle

\IEEEdisplaynontitleabstractindextext

\IEEEpeerreviewmaketitle

\input{Secs/1_Introduction.tex}

\input{Secs/2_Background.tex}

\input{Secs/3_Meta_Architecture}
\input{Secs/4_Image_Scene}
\input{Secs/5_Video_Scene}

\input{Secs/6_3D_Scene}

\input{Secs/7_GREx.tex}
\input{Secs/8_Applications.tex}
\input{Secs/10_Conclusion.tex}

\input{Secs/9_Performance_Comparison.tex}

\ifCLASSOPTIONcaptionsoff
  \newpage
\fi

{
\bibliographystyle{IEEEtran}
\bibliography{IEEEabrv,egbib}
}

\end{document}

%% file: Secs/0_Abstract.tex
\IEEEtitleabstractindextext{
\begin{abstract}
Multimodal referring segmentation aims to segment target objects in visual scenes, such as images, videos, and 3D scenes, based on referring expressions in text or audio format. This task plays a crucial role in practical applications requiring accurate object perception based on user instructions. Over the past decade, it has gained significant attention in the multimodal community, driven by advances in convolutional neural networks, transformers, and large language models, all of which have substantially improved multimodal perception capabilities. This paper provides a comprehensive survey of multimodal referring segmentation. We begin by introducing this field's background, including problem definitions and commonly used datasets. Next, we summarize a unified meta architecture for referring segmentation and review representative methods across three primary visual scenes, including images, videos, and 3D scenes. We further discuss Generalized Referring Expression (GREx) methods to address the challenges of real-world complexity, along with related tasks and practical applications. Extensive performance comparisons on standard benchmarks are also provided. We continually track related works at \url{https://github.com/henghuiding/Awesome-Multimodal-Referring-Segmentation}.
\end{abstract}

}

%% file: Secs/1_Introduction.tex
\section{Introduction}
\label{sec:introduction}

\begin{figure}[t]
	\centering
	\includegraphics[width=1\linewidth]{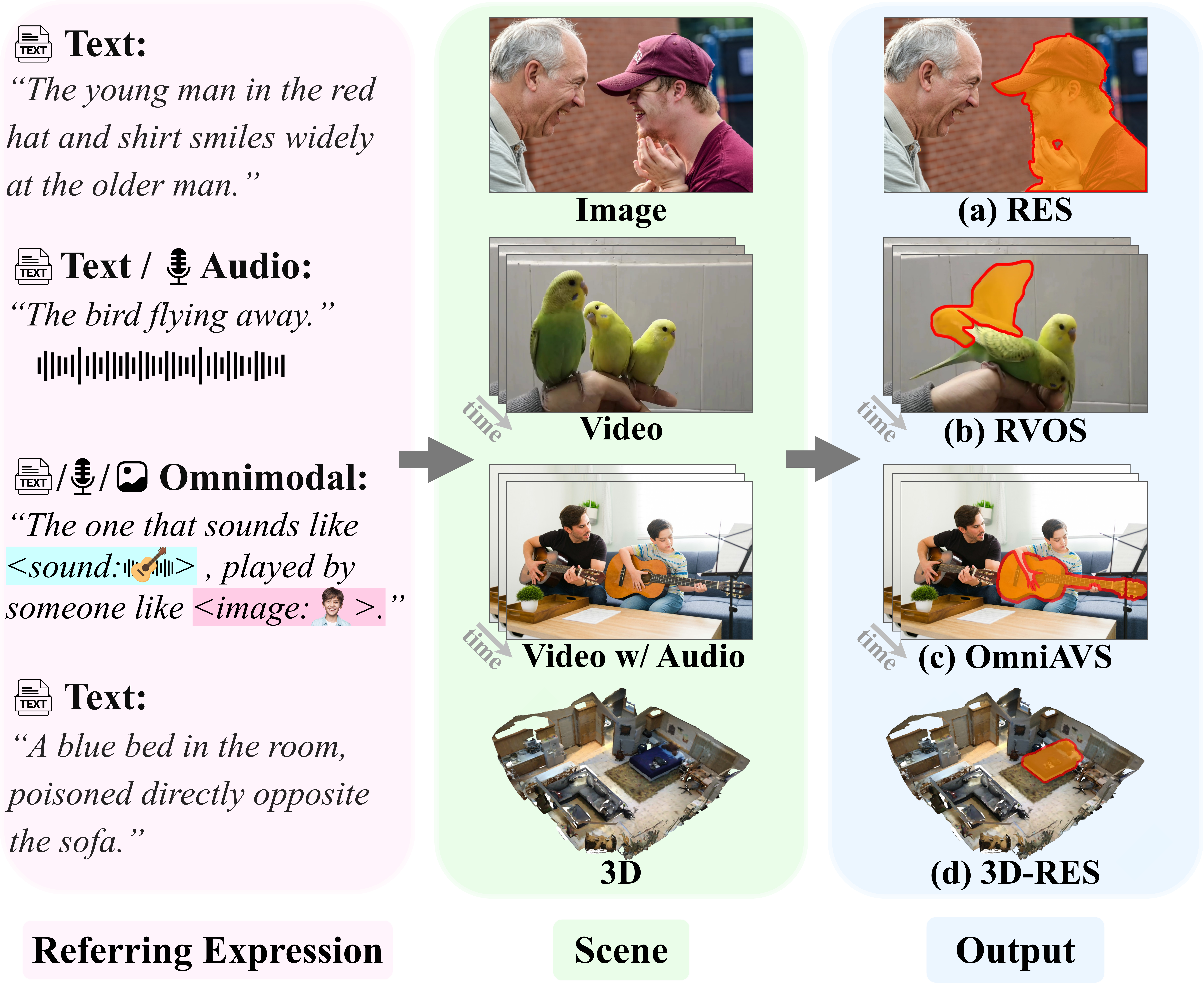}
	\caption{\textbf{Multimodal Referring Segmentation.}}
 	\label{fig:task_overview}
    \vspace{-1\baselineskip}
\end{figure}

\begin{figure}[t]
	\centering
        \begin{overpic}[width=1\linewidth]{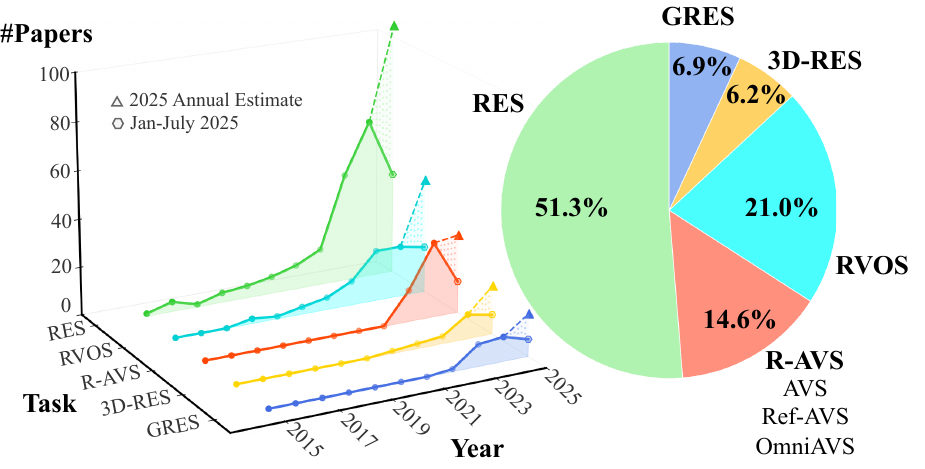}
            \put(8.7,121.2){\fontsize{8}{12}\selectfont \textbf{(\S~\!\ref{sec:introduction})}}
            \put(28.1,121.2){\fontsize{8}{12}\selectfont \textbf{(\S~\!\ref{sec:background})}}
            \put(50.3,121.2){\fontsize{8}{12}\selectfont
            \textbf{(\S~\!\ref{sec:meta_architecture})}}
            \put(72.8,121.2){\fontsize{8}{12}\selectfont 
            \textbf{(\S~\!\ref{sec:res})}}
            \put(92.0,121.2){\fontsize{8}{12}\selectfont
            \textbf{(\S~\!\ref{sec:rvos})}}
            \put(109.8,121.2){\fontsize{8}{12}\selectfont
            \textbf{(\S~\!\ref{sec:r-avs})}}
            \put(127.1,121.2){\fontsize{8}{12}\selectfont
            \textbf{(\S~\!\ref{sec:3d-res})}}
            \put(146.2,121.2){\fontsize{8}{12}\selectfont
            \textbf{(\S~\!\ref{sec:grex})}}
            \put(165.5,121.2){\fontsize{8}{12}\selectfont 
            \textbf{(\S~\!\ref{sec:related})}}
            \put(184.3,121.2){\fontsize{8}{12}\selectfont
            \textbf{(\S~\!\ref{sec:conclusion})}}

                \put(9.3,108.7){\fontsize{7}{12}\selectfont (\S~\!\ref{sec2.1})}
                \put(9.3,101.2){\fontsize{7}{12}\selectfont   (\S~\!\ref{sec2.2})}

                \put(38.35,108.7){\fontsize{7}{12}\selectfont (\S~\!\ref{sec3.1})}
                \put(38.35,101.2){\fontsize{7}{12}\selectfont   (\S~\!\ref{sec3.2})}
                \put(38.35,93.9){\fontsize{7}{12}\selectfont   (\S~\!\ref{sec3.3})}
                \put(38.35,86.5){\fontsize{7}{12}\selectfont   (\S~\!\ref{sec3.4})}
                \put(38.35,79){\fontsize{7}{12}\selectfont   (\S~\!\ref{sec3.5})}
                \put(38.35,71.8){\fontsize{7}{12}\selectfont   (\S~\!\ref{sec3.6})}

                \put(72.05,108.7){\fontsize{7}{12}\selectfont (\S~\!\ref{sec4.1})}
                \put(72.05,101.2){\fontsize{7}{12}\selectfont   (\S~\!\ref{sec4.2})}
                \put(72.05,93.9){\fontsize{7}{12}\selectfont   (\S~\!\ref{sec4.3})}
                \put(72.05,86.5){\fontsize{7}{12}\selectfont   (\S~\!\ref{sec4.4})}
                \put(72.05,79){\fontsize{7}{12}\selectfont   (\S~\!\ref{sec4.5})}
                \put(72.05,71.8){\fontsize{7}{12}\selectfont   (\S~\!\ref{sec4.6})}
                \put(72.05,64.2){\fontsize{7}{12}\selectfont   (\S~\!\ref{sec4.7})}

                \put(107.45,108.7){\fontsize{7}{12}\selectfont  (\S~\!\ref{sec5.1})}
                \put(107.45,101.2){\fontsize{7}{12}\selectfont   (\S~\!\ref{sec5.2})}
                \put(107.45,93.9){\fontsize{7}{12}\selectfont   (\S~\!\ref{sec5.3})}
                \put(107.45,86.5){\fontsize{7}{12}\selectfont   (\S~\!\ref{sec5.4})}

                \put(143.85,108.7){\fontsize{7}{12}\selectfont  (\S~\!\ref{sec6.1})}
                \put(143.85,99.4){\fontsize{7}{12}\selectfont   (\S~\!\ref{sec6.2})}
                \put(143.85,90){\fontsize{7}{12}\selectfont   (\S~\!\ref{sec6.3})}

                \put(180,106.1){\fontsize{7}{12}\selectfont   (\S~\!\ref{sec7.1})}
                \put(180,96.1){\fontsize{7}{12}\selectfont   (\S~\!\ref{sec7.2})}

                \put(142,76.6){\fontsize{7}{12}\selectfont   \textbf{(\S~\!{\hyperref[sec:performance_comparison]{Appendix}})}}
                
            \put(114.7,70.5){\fontsize{7.5}{12}\selectfont (\S~\!\ref{RESBench})}
            \put(145.4,70.5){\fontsize{7.5}{12}\selectfont (\S~\!\ref{RVOSBench})}
            \put(176.1,70.5){\fontsize{7.5}{12}\selectfont (\S~\!\ref{RAVSBench})}
            \put(114.7,63.05){\fontsize{7.5}{12}\selectfont (\S~\!\ref{3DRESBench})}
            \put(145.4,63.05){\fontsize{7.5}{12}\selectfont (\S~\!\ref{GRExBench})}
            \put(176.1,63.05){\fontsize{7.5}{12}\selectfont (\S~\!\ref{ReasonBench})}
        \end{overpic}
         \vspace{-6mm}
	\caption{Publication statistics of multimodal referring segmentation papers in top conferences/journals of computer vision, machine learning, and artificial intelligence, collected up to July 2025.
    }
  	\label{fig:curve}
    \vspace{-0.8\baselineskip}
\end{figure}

\IEEEPARstart{M}{ultimodal} referring segmentation~\cite{MeViS,MeViSv2,gres,vlt,Ref-LERF,lstm-cnn,vlticcv,ReferItGame,segpoint} aims to segment the target object in a visual scene of image~\cite{gres,vlt,vlticcv}, video~\cite{MeViS,MeViSv2,OmniAVS}, or 3D~\cite{3d-gres,Ref-LERF,segpoint} according to a referring expression, such as free-form text or audio. For example, as shown in \cref{fig:task_overview}(b), given the text referring expression \textit{``The bird flying away''}, the model is expected to segment and track the described target object in the video. This task presents a fundamental and challenging problem in multimodal understanding and supports a wide range of practical applications, such as image/video editing~\cite{rie,image_editing_survey}, robotics~\cite{fang2020graspnet}, autonomous driving~\cite{echotrack}, \etc. Because of its significant potential in practical applications, multimodal referring segmentation has received growing attention in recent years, as shown in \cref{fig:curve}.

\begin{figure*}[t]
	\includegraphics[width=1\linewidth]{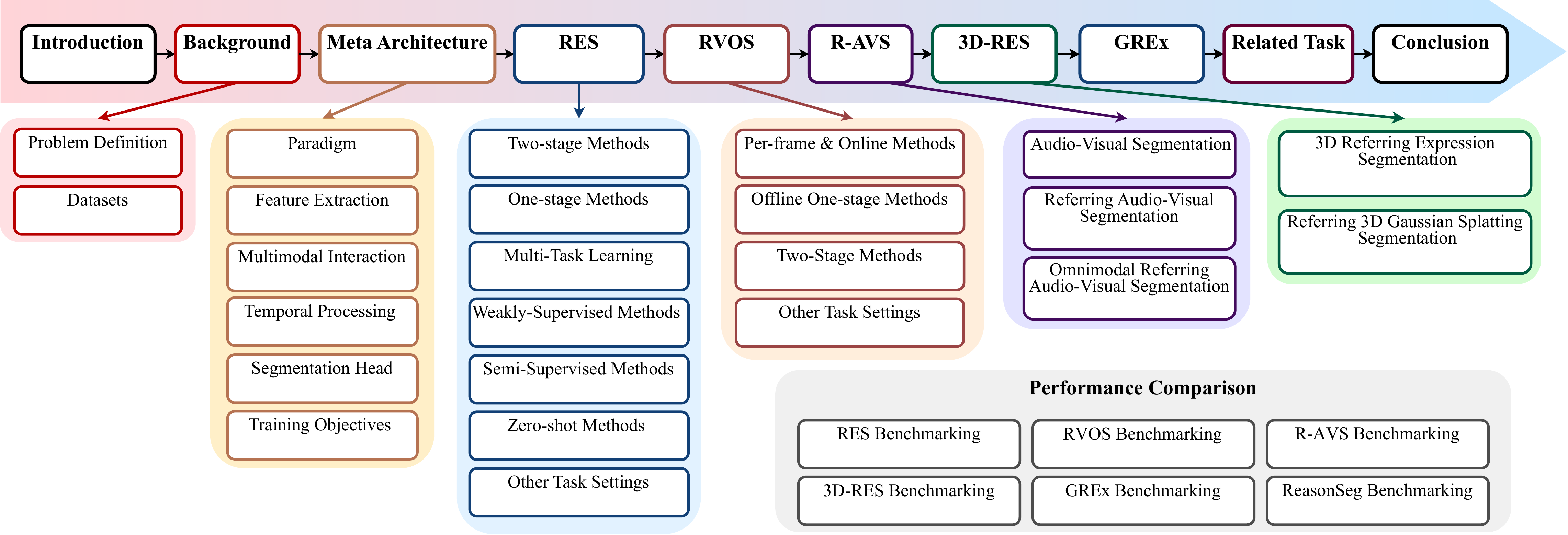}
        \vspace{-6mm}
	\caption{\textbf{Overview of this survey.} Different colors represent specific sections. Best viewed in color.
	}
	\label{fig:overview}
    \vspace{-3mm}
\end{figure*}

Segmentation~\cite{fcn,maskrcnn,CCL} is one of the fundamental tasks in computer vision, forming the basis for many visual understanding tasks and applications~\cite{li2024transformer}. Classic segmentation methods, \eg, semantic segmentation~\cite{fcn} and instance segmentation~\cite{maskrcnn}, typically segment the given visual scenes into a set of predefined categories. Although open-vocabulary segmentation~\cite{Survey_OpenVocabulary} expands the category coverage, it remains reliant on explicit category names, \eg, \textit{person} and \textit{car}. Different from these classical segmentation tasks, referring segmentation enables more flexible and user-friendly segmentation by leveraging free-form referring expressions to identify specific target objects within a scene. A referring expression is a human-understandable linguistic construct used to describe an object in any way that uniquely and unambiguously identifies it. Such expressions are not limited to naming object categories. They may refer to the target object’s position, visual attributes, motion, or relationships with other objects. As long as the expression leads to an unambiguous identification of the target, any descriptive strategy is considered valid. This high degree of expressive freedom introduces considerable challenges for fine-grained multimodal understanding and alignment. It also raises requirements for model robustness against diverse expression styles and linguistic-visual variations. Depending on the modality of the referring signal (\eg, text or audio) and the type of visual scene (\eg, image, video, auditory video, or 3D), referring segmentation can be further categorized into different tasks, as shown in \cref{fig:task_overview}.

\begin{figure}[ht]
    \centering
    \begin{overpic}[width=1\linewidth]{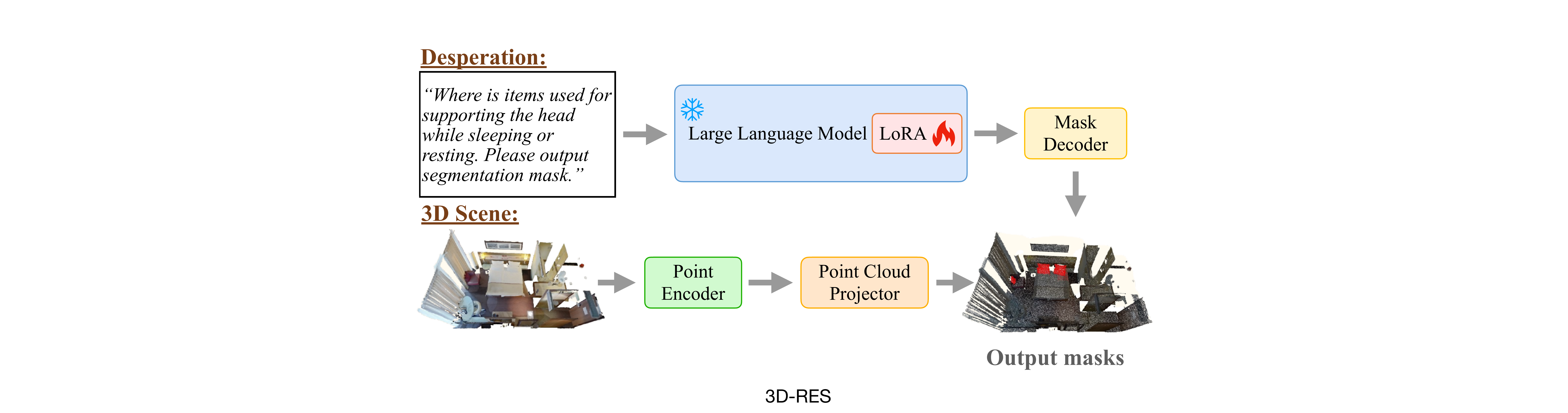}
        \put(14.6,5.4){\tiny\cite{lisa}}
        \put(28.3,5.4){\tiny\cite{segpoint}}
        \put(44.7,5.4){\tiny\cite{dshmp}}

        \put(21.6,9.8){\tiny\cite{vlt}}
        \put(35.7,9.8){\tiny\cite{Ref-LERF}}
        \put(52.7,9.8){\tiny\cite{MeViS}}
        \put(70,9.8){\tiny\cite{ref-avs}}

        \put(44.7,18.93){\tiny\cite{3d-avs}}
        \put(59.5,18.93){\tiny\cite{MeViSv2}}
        \put(79,18.93){\tiny\cite{teso}}

        \put(85.2,26.7){\tiny\cite{OmniAVS}}
    \end{overpic}
    \vspace{-6mm}
    \caption{\textbf{Overview of Multimodal Referring Segmentation tasks.} Representative works are cited in the top-left corners of each task.}
    \label{fig:taskoverview}
    \vspace{-3mm}
\end{figure}
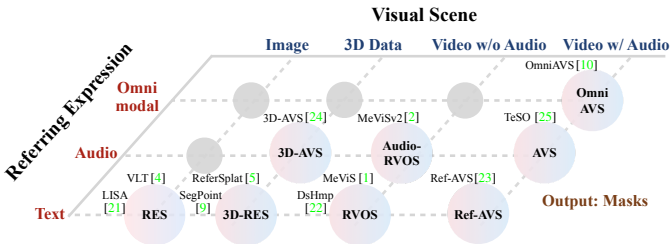

Despite the inherent homogeneity shared across different referring segmentation tasks, most existing surveys~\cite{rec_survey,3d_rec_survey,reasoningseg_survey,wei2022learning,rec2024} remain limited in scope, often concentrating on isolated modalities or specific tasks. For example, a recent survey~\cite{survey_res} focuses exclusively on referring expression segmentation within 2D images, while neglecting extensions to video and 3D scenes. As a result, a critical gap remains in the literature due to the absence of a comprehensive survey that systematically covers the diverse task formulations, input modalities, and challenges within referring segmentation. Addressing this gap is essential for fostering a deeper understanding of the field and for advancing the development of generalizable and multimodal solutions.

To this end, we conduct a comprehensive review of over 600 papers in the field of multimodal referring segmentation. This survey seeks to unify diverse referring modalities across various visual scenes. Our goal is to offer a cohesive and structured understanding of the field to enhance accessibility and facilitate cross-task insights. In addition, we highlight practical applications of referring expression techniques, demonstrating their transformative potential in emerging domains such as embodied AI.

\noindent$\bullet$ \textbf{Scope.}
This survey focuses on recent advances in referring segmentation across three major visual scenes: image-based, video-based (including salient and auditory videos), and 3D-based scenarios, along with three primary referring modalities: text, audio, and omnimodal, as shown in \cref{fig:taskoverview}. It primarily reviews deep learning-based methods, highlighting influential works published in top-tier conferences and journals, along with recent preprints that reflect emerging trends and future directions.

\noindent$\bullet$~\textbf{Organization.}
An overview of the survey is shown in \cref{fig:overview}. We begin with background on problem definitions and datasets in \cref{sec:background}, followed by a unified meta architecture in \cref{sec:meta_architecture} that spans various referring segmentation tasks. Based on this framework, representative methods across image, video, and 3D scenes are systematically reviewed in \cref{sec:res} to \cref{sec:3d-res}. Considering the real-world complexity, we further discuss Generalized Referring Expression (GREx) in \cref{sec:grex}. Related tasks and applications are explored in \cref{sec:related}, followed by the conclusion and discussion in \cref{sec:conclusion}. Benchmark results are provided in the Appendix.

%% file: Secs/2_Background.tex
\section{Background}
\label{sec:background}

\subsection{Problem Definition}
\label{sec2.1}

\begin{figure*}[t]
    \centering
    \includegraphics[width=1\linewidth]{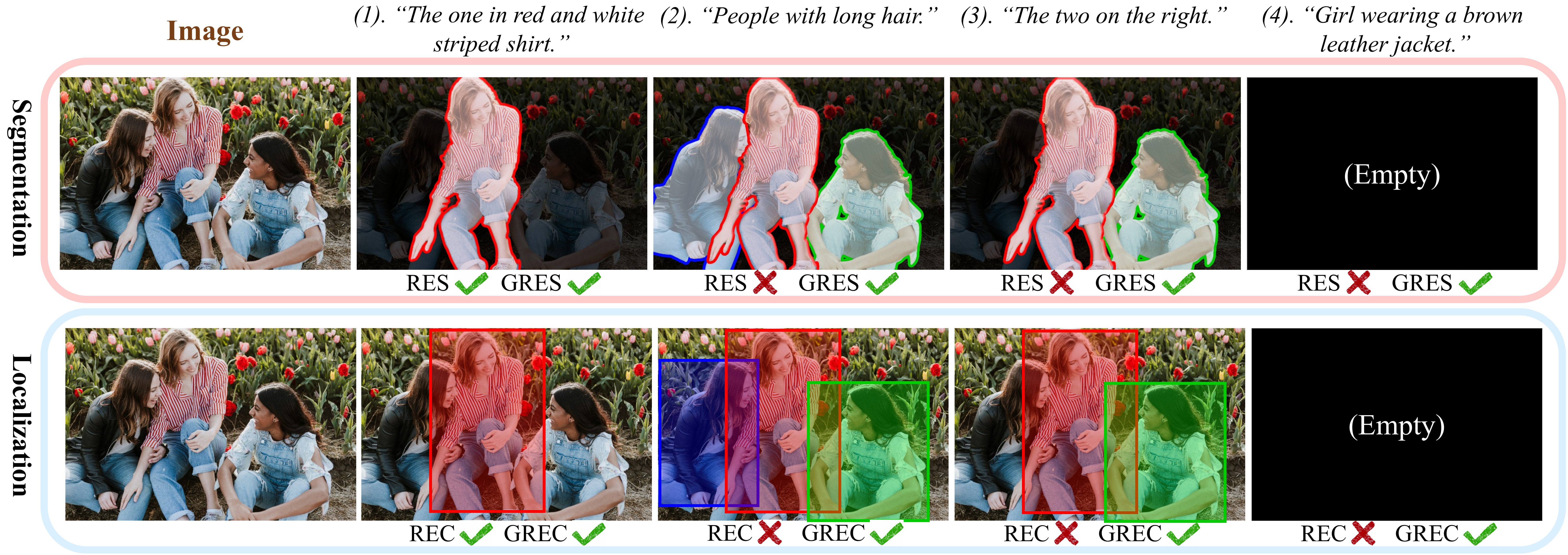}
    \vspace{-6.6mm}
    \caption{Classic Referring Expression 
     Segmentation (RES) and Comprehension (REC) handle expressions that refer to a single target object, as shown in example (1). In contrast, Generalized Referring Expression 
    Segmentation (GRES)~\cite{gres} and Comprehension (GREC)~\cite{grec,hiea2g} support expressions referring to any number of target objects, including multi-target expressions like (2) and (3), as well as no-target expressions such as (4), thereby enhancing their applicability in complex and diverse real-world scenarios.}
    \label{fig:GREx}
    \vspace{-3mm}
\end{figure*}

\subsubsection{A Unified Formulation}

The primary objective of this survey is to provide a systematic investigation of the family of referring segmentation tasks. To this end, we propose a unified formulation that generalizes across different task variants within this domain.
Specifically, let $\mathcal{X}$ and $\mathcal{Y}$ denote the input and output spaces, respectively. The objective is to learn an optimal mapping function $f$ defined as:
\begin{equation}
\label{Eq:unifiedFormulation}
f : \mathcal{X} \mapsto \mathcal{Y}, \quad \text{where } \mathcal{X} = \mathcal{V} \times {E},
\end{equation}
where $f$ is typically instantiated as a neural network. The input space $\mathcal{X}$ consists of two components: the visual input $\mathcal{V}$ (\eg, image, silent video, auditory video, or 3D data), and the referring signal ${E}$ (\eg, text, audio, etc.) that specifies the target object(s) of interest. The output space $\mathcal{Y}$ consists of segmentation mask(s) of the referred entities within $\mathcal{V}$.
Based on this unified formulation, we construct a comprehensive taxonomy of referring segmentation tasks and formally define each task in the following sections.

\subsubsection{Image Scene}
\noindent$\bullet$ \textit{\textbf{Referring Expression Segmentation (RES).}}
RES, also known as Referring Image Segmentation (RIS), aims to segment the target object in a given image ${I} \in \mathbb{R}^{H \times W \times 3}$ according to a natural language referring expression ${E}$. The output is a binary mask ${M} \in \{0,1\}^{H \times W}$ that precisely delineates the object referred to by ${E}$. Compared to semantic or instance segmentation tasks that rely on predefined object categories, RES introduces new challenges in comprehending the linguistic content of ${E}$ and reasoning about the complex visual relationships among objects in the scene, such as spatial configurations, attributes, and interactions.

\noindent$\bullet$ \textit{\textbf{Reasoning Segmentation.}}  Reasoning Segmentation is a special case of RES where the referring expression requires indirect reasoning rather than explicit object descriptions, \eg, “\textit{the food with the most Vitamin C}”. Recent advances in LLMs/MLLMs have made it feasible to handle such expressions by leveraging their strong reasoning and commonsense capabilities. 

\begin{table*}
\centering
\caption{\textbf{Representative Datasets for Referring Segmentation and Their Characteristics}}
\vspace{-3mm}
\begin{threeparttable}
\fontsize{10pt}{12pt}\selectfont
\resizebox{0.99\textwidth}{!}{
\setlength\tabcolsep{6pt}
\renewcommand\arraystretch{1.0}
\begin{tabular}{r||c|c|c|p{15cm}}
\specialrule{.1em}{.05em}{.05em} 
\rowcolor{mygray}
\textbf{Dataset} & \textbf{\#Imgs/Vids/3D Scenes} & \textbf{\#Expressions} & \textbf{\#Objects} & \textbf{Characterization} \\
\hline
\hline
\multicolumn{2}{l}{\textbf{\textit{Image Scene}}} \\
   \rowcolor{cyan!07}
ReferItGame~\cite{ReferItGame} & 19,894 & 130,525 & 96,654 &Focusing on real-world expressions but is limited by simpler descriptions. \\ 
RefCOCO~\cite{RefCOCO&RefCOCO+&RefCOCO(g)} & 19,994 & 142,209 & 50,000 & Allowing both location- and appearance-based references in images. \\
   \rowcolor{cyan!07}
RefCOCO+~\cite{RefCOCO&RefCOCO+&RefCOCO(g)} & 19,992 & 141,564 & 49,856 & Focusing on appearance-based expressions without location-based descriptions. \\
RefCOCOg~\cite{RefCOCO&RefCOCO+&RefCOCO(g)} & 25,799 & 95,010 & 49,822 & Including longer and more complex expressions without restrictions on location. \\ 
   \rowcolor{cyan!07}
PhraseCut~\cite{phrasecut} & 77,262 & 345,486 & - & Focusing on fine-grained expressions covering categories, attributes, and relationships in diverse scenes.
 \\
ReasonSeg~\cite{lisa} & 1,218 & - & - & Focusing on implicit, reasoning-based expressions requiring world knowledge and complex understanding.
 \\ 
\hline
\multicolumn{2}{l}{\textbf{\textit{Video Scene}}} \\
   \rowcolor{cyan!07}
MeViS~\cite{MeViS} & 2,006 & 28,570 & 8,171 &Focusing on motion attributes and enabling multi-object expressions for enhanced RVOS challenges. \\
MeViSv2~\cite{MeViSv2} & 2,006 & 33,072 & 8,171 &Based on MeViS, further adding no-target and motion reasoning expressions, adding audio expressions. \\
\rowcolor{cyan!07}
Ref-DAVIS$_{16}$~\cite{Refer-DAVIS} & 50 & 100 & 50 &Containing single-object video sequences annotated with referring expressions. \\

Ref-DAVIS$_{17}$~\cite{Refer-DAVIS} & 90 & 1,544 & 205 & Presenting more challenging multi-object scenarios involving occlusions and distractors. \\
   \rowcolor{cyan!07}
A2D Sentences~\cite{A2D&J-HMDB} & 3,782 & 6,656 & 4,825& Enriching A2D~\cite{A2D} with elaborate natural language descriptions for actor and action segmentation. \\
J-HMDB Sentences~\cite{A2D&J-HMDB} & 928 & 928 & 928 &Supplementing J-HMDB~\
\cite{J-HMDB} with natural language descriptions, mainly concerning human actions. \\
   \rowcolor{cyan!07}
Refer-Youtube-VOS~\cite{Refer-Youtube-VOS} & 3,975 & 27,899 & 7,451 & Offering pixel-level RVOS across multiple object categories. \\
ReVOS~\cite{visa} & 1,042 & 35,074 & -  &Focusing on implicit, complex text queries requiring world knowledge and video context for segmentation.\\
   \rowcolor{cyan!07}
ReasonVOS~\cite{videolisa}& 91 & 458 & - &Evaluating models' reasoning ability using complex language queries and world knowledge.
\\
\hline
\multicolumn{2}{l}{\textbf{\textit{Audio-Video Scene}}} \\
   \rowcolor{cyan!07}
AVSBench-S4~\cite{avsbench} & 4,932 & - & - & Focusing on individual sound-making objects in semi-supervised single sound source AVS tasks. \\
AVSBench-MS3~\cite{avsbench} & 424 & - & - & Focusing on concurrent sound sources in fully supervised multiple sound source AVS scenarios. \\
   \rowcolor{cyan!07}
AVSBench-Semantic~\cite{avsbench_semantic} & 12,356 & - & - & Offering semantic annotations for fully supervised audio-visual semantic segmentation. \\
Ref-AVS~\cite{ref-avs} & 4,002 & 20,261 & 6,888 &Providing pixel-level annotations for objects described in corresponding multimodal-cue expressions. \\
   \rowcolor{cyan!07}
OmniAVS~\cite{OmniAVS}  & 2,098 & 59,458 & 4,262 & Providing 8 different omnimodal expressions flexibly consisting of text, speech, sound, and image. \\
\hline  
\multicolumn{2}{l}{\textbf{\textit{3D Scene}}} \\
   \rowcolor{cyan!07}
ScanRefer~\cite{scanrefer} & 800 & 51,583 & 11,046  & Pioneering the first large-scale dataset for object localization via natural language expressions. \\
Nr3D~\cite{referit3d} & 707 & 41,503 & 5,878 & Providing human-annotated descriptions for precise 3D object localization in real-world 
scenes. \\
   \rowcolor{cyan!07}
Sr3D~\cite{referit3d} & 1,273 & 83,572 & 11,375 &Offering synthetically generated expressions with simplified language patterns. \\
Instruct3D~\cite{segpoint} & 280 & 2,565 & - & Supporting 3D instruction segmentation from complex texts, including both multi- and zero-target scenes.\\
\rowcolor{cyan!07}
Ref-LERF~\cite{Ref-LERF}&4 & 295&59 &Focusing on spatial relationships for referring 3D gaussian splatting segmentation.\\
\hline
\multicolumn{2}{l}{\textbf{\textit{GREx}}} \\
   \rowcolor{cyan!07}
gRefCOCO~\cite{gres} & 19,994 & 278,232 & 60,287 & Extending RefCOCO~\cite{RefCOCO&RefCOCO+&RefCOCO(g)} by supporting multi-target and no-target expressions. \\

Ref-ZOM~\cite{dmmi} & 55,078 & 90,199 & 74,942 & Supporting GRES tasks with annotations built on COCO dataset. \\
   \rowcolor{cyan!07}
Multi3DRefer~\cite{multi3drefer} & 800 & 61,926 & 11,609 & Extending
ScanRefer~\cite{scanrefer} with zero/single/multiple target descriptions for supporting 3D-GREC task. \\

Multi3DRes~\cite{3d-gres} & 800 & 61,926 & 11,609 & Adapting Multi3DRefer~\cite{multi3drefer} with instance masks to support 3D-GRES task. \\

\specialrule{.1em}{.05em}{.05em} 
\end{tabular}
}
\end{threeparttable}
    \label{tab:datasets}
\vspace{-3.6mm}
\end{table*}

\subsubsection{Video Scene} \label{sec:def-video}
\noindent$\bullet$ \textit{\textbf{Referring Video Object Segmentation (RVOS).}}
RVOS extends referring segmentation to the video domain.
Given a video $\textbf{V} = \{V_t\}_{t=1}^T$ with $T$ frames, where each frame $V_t \in \mathbb{R}^{H \times W \times 3}$, and a natural language referring expression $E$, the goal of RVOS is to generate a sequence of binary masks $\textbf{M} = \{M_t\}_{t=1}^T$, where each $M_t \in \{0,1\}^{H \times W}$ denotes the pixel-level segmentation of the referred object in frame $V_t$. Compared to RES, RVOS introduces additional challenges, such as maintaining temporal consistency across frames, dealing with occlusions and appearance variations, and tracking the referred object despite partial or full occlusion.

\noindent$\bullet$ \textit{\textbf{Audio-Visual Segmentation (AVS).}}
AVS aims to segment the sound-emitting objects throughout an auditory video. Given an auditory video $\{\textbf{V}, \textbf{A}\}$, where $\textbf{V} = \{V_t\}_{t=1}^{T}$ denotes $T$ visual frames and $\textbf{A} = \{A_t\}_{t=1}^{T}$ is the corresponding audio stream, the goal of AVS is to predict binary masks $\textbf{M} = \{M_t\}_{t=1}^{T}$, with each $M_t \in {0,1}^{H \times W}$ highlighting the regions in $V_t$ associated with the sound source in $A_t$. AVS is a special case of referring video segmentation, where the query is implicitly defined as ``\textit{segment the sound-emitting objects in the video}.''

\noindent$\bullet$ \textit{\textbf{Referring Audio-Visual Segmentation (Ref-AVS).}}
Ref-AVS aims to segment target objects in an auditory video $\{\textbf{V}, \textbf{A}\}$ according to a text referring expression $E$. The desired output is a sequence of binary masks $\textbf{M} = \{M_t\}_{t=1}^T$, where each $M_t \in \{0,1\}^{H \times W}$ is the pixel-level segmentation of the referred object in frame $V_t$. Ref-AVS enables handling scenarios that are difficult to address in AVS and RVOS, \eg, \textit{``segment the person singing bass in the a cappella group"}. This poses the importance of leveraging multi-modal cues to guide visual segmentation.

\noindent$\bullet$ \textit{\textbf{Omnimodal Referring Audio-Visual Segmentation (OmniAVS).}}
{OmniAVS} aims to segment specific target objects in an auditory video $\{\textbf{V}, \textbf{A}\}$ according to a multimodal expression $E$ that flexibly combines text, speech, sound, and visual cues. The desired output is a sequence of binary masks $\textbf{M} = \{M_t\}_{t=1}^T$, where each $M_t \in \{0,1\}^{H \times W}$ denotes the pixel-level segmentation of the referred object in frame $V_t$. The ability to handle diverse multimodal expressions makes OmniAVS both practical for real-world applications and well-suited for advancing omnimodal models with fine-grained perceptual capabilities.

\subsubsection{3D Scene}
\noindent$\bullet$ \textit{\textbf{3D Referring Expression Segmentation (3D-RES).}}~3D-RES aims to segment the target object within a 3D scene based on a referring expression $E$. Given a 3D point cloud consisting of $N$ points, denoted as $\mathbf{P}=\{P_i\}_{i=1}^N$, and a referring expression $E$, the objective is to produce a binary mask $M\in \{0,1\}^{N}$ that identifies the subset of points corresponding to the object referred to by $E$. Compared to 2D RES on structured image grids, 3D-RES involves segmenting target points in unordered, irregular, and sparse point clouds, requiring both effective language–vision alignment and a deep understanding of geometric structures.

\noindent$\bullet$ \textit{\textbf{Referring 3D Gaussian Splatting Segmentation (R3DGS).}} Given a scene with $S$ multi-view RGB images $\mathbf{I} = \{I_i\}_{i=1}^S$ and $L$ corresponding referring expressions $\mathbf{E} = \{E_l\}_{l=1}^L$ during training, R3DGS aims to segment the target object in a {novel-view} image $I \in \mathbb{R}^{H \times W \times 3}$ of the scene based on a given expression $E$. The output is a binary mask $M \in \{0,1\}^{H \times W}$ that delineates the target object, potentially under occlusion. Unlike conventional 2D referring segmentation, R3DGS focuses on 3D scene reconstructed from multi-view images. Compared to existing 3D referring tasks that rely on point clouds and 3D mask supervision, R3DGS learns from 2D images without requiring explicit 3D annotations, offering a more scalable and annotation-efficient paradigm.

\subsubsection{GREx}

\noindent$\bullet$ \textit{\textbf{Generalized Referring Expression Segmentation (GRES).}}
As shown in \cref{fig:GREx}, GRES~\cite{gres} extends the scope of referring segmentation by allowing expressions to refer to any number of target objects. Given a visual input $\mathcal{V}$ and a referring expression $E$, GRES aims to predict a binary mask $M$ covering all relevant pixels or points corresponding to the described target object(s). Unlike conventional settings that focus solely on single object, GRES supports single-target, multi-target, and no-target expressions, improving models' adaptability in real-world scenarios. This generalization introduces new challenges, particularly in achieving precise alignment between different modalities when dealing with ambiguous, descriptive, or compositional expressions.

\noindent$\bullet$ \textit{\textbf{Generalized Referring Expression Comprehension (GREC).}}
Parallel to GRES, GREC~\cite{grec} is introduced to expand the scope of the classic REC task, see \cref{fig:GREx}. In contrast to classic REC that generates a single bounding box for a sentence, GREC pursues the generation of a collection of bounding boxes within the input $\mathcal{V}$, denoted as $\mathbf{B}=\{B_i\}$, wherein each bounding box $B_i\in\mathbb{R}^4$ encloses an object among the entirety of target objects indicated by the  expression $E$. The number of bounding boxes may vary from 0 to multiple, depending on the given expression.

Beyond image scenes, GRES and GREC can be applied to video and 3D scenes, leading to task variants such as Video-GRES \cite{MeViS}, AV-GRES~\cite{OmniAVS}, and 3D-GRES \cite{3d-gres}, expanding the applicability of referring segmentation to real-world scenarios.

\subsection{Datasets} 
\label{sec2.2}
\begin{figure}[t]
    \centering
    \includegraphics[width=1\linewidth]{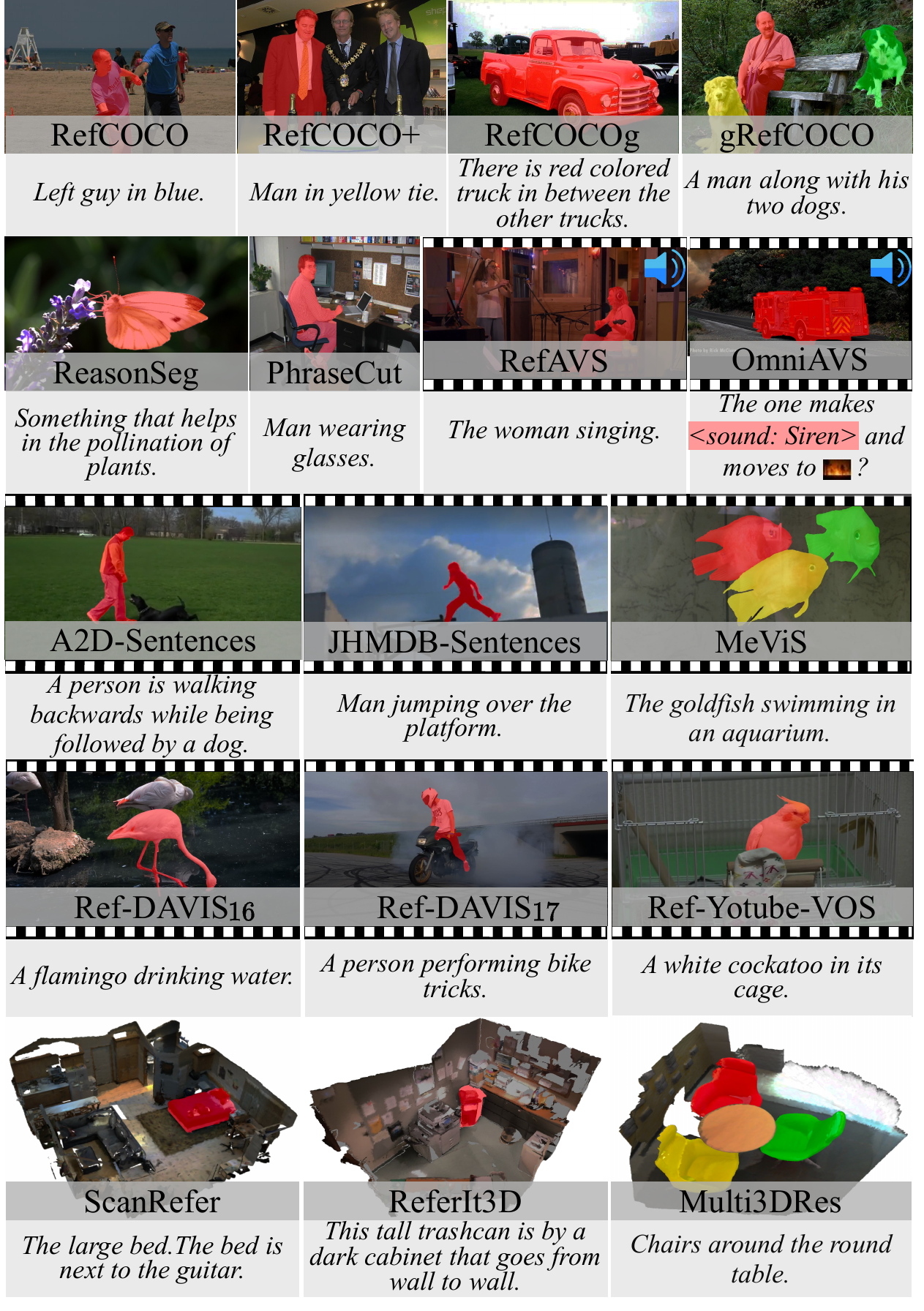}
    \vspace{-7mm}
    \caption{Examples from 17 commonly used referring segmentation datasets, including image, video, and 3D scene data.
    }
    \label{fig:dataset-segmentation}
    \vspace{-3mm}
\end{figure}

We briefly introduce commonly used referring segmentation datasets. \Cref{tab:datasets} lists more datasets and summarizes their key characteristics, and \cref{fig:dataset-segmentation} presents some representative examples.

\vspace{-1.6mm}
\subsubsection{Image Scene}
\noindent$\bullet$~\textbf{ReferItGame}~\cite{ReferItGame} contains 130,525 referring expressions for 96,654 objects across 19,894 images. As the first large-scale dataset for referring expression understanding, it is a valuable resource despite its relatively simple language expressions.

\noindent$\bullet$~\textbf{RefCOCO/+/g}~\cite{RefCOCO&RefCOCO+&RefCOCO(g)} contains 142,209 expressions for 50,000 objects (RefCOCO), 141,564 expressions for 49,856 objects (RefCOCO+), and 104,560 expressions for 54,822 objects (RefCOCOg). These datasets include over 19,000 images for RefCOCO and RefCOCO+, and 26,711 images for RefCOCOg. RefCOCO allows the use of spatial clues. RefCOCO+ limits the use of such terms, focusing more on the visual attributes. RefCOCOg contains longer and more complex expressions.

\noindent$\bullet$~\textbf{ReasonSeg}~\cite{lisa} sets a benchmark for the reasoning segmentation. It includes 1,218 image-instruction-mask samples with implicit text instructions and target masks.

Beyond the above, several other datasets~\cite{CLEVR-Ref+,phrasecut,mres,aeroreformer,UniRES++,GroundingSuite,RVTBench,SynRES} contribute unique challenges to RES research.

\vspace{-1.mm}
\subsubsection{Video Scene}
\noindent$\bullet$~\textbf{MeViS}~\cite{MeViS} contains 2,006 videos with 28,570 motion-based referring expressions, emphasizing temporal motion for segmentation and excelling in tasks requiring long-term motion comprehension. MeViS has 4.28 objects per video on average.

\noindent$\bullet$~\textbf{MeViSv2}~\cite{MeViSv2} extends MeViS~\cite{MeViS} with 4,502 additional motion reasoning and no-target expressions, and further provides audio expressions to support audio-based RVOS.

\noindent$\bullet$~\textbf{Ref-DAVIS}~\cite{Refer-DAVIS} comprises 1,200 referring expressions annotated for over 400 objects across 150 videos with around 10K frames, built upon the DAVIS$_{16}$ and DAVIS$_{17}$~\cite{davis16} datasets.

\noindent$\bullet$~\textbf{Ref-YouTube-VOS}~\cite{Refer-Youtube-VOS} contains 3,978 videos and 15,009 expressions with pixel-level annotations. Two types of expressions are provided: 1) full-video expressions containing both static appearance and dynamic temporal clues, and 2) first-frame expressions only using static appearance clues of the first frame. 

\noindent$\bullet$~\textbf{A2D Sentences}~\cite{A2D&J-HMDB} contains 6,656 sentences describing actions performed by actors across 3,782 videos, offering actor-action segmentation for videos. It has 1.28 objects per video on average.

\noindent$\bullet$~\textbf{J-HMDB Sentences}~\cite{A2D&J-HMDB} provides 928 referring expressions for human actions in 928 videos of J-HMDB dataset, supporting action localization and segmentation.

\vspace{-1.16mm}
\subsubsection{Auditory Video Scenes}
\noindent$\bullet$~\textbf{AVSBench-object}~\cite{avsbench} 
provides two settings: Single-source (S4) includes 4,932 videos from 23 categories with sparse supervision (only the first frame labeled), while Multi-source (MS3) contains 424 fully annotated videos with multiple sound sources. 

\noindent$\bullet$~\textbf{Ref-AVS}~\cite{ref-avs} provides 20,261 text expressions for 6,888 objects in 4002 auditory videos. The text expressions adopt multimodal cues to describe objects in audio-visual scenes.

\noindent$\bullet$~\textbf{OmniAVS}~\cite{OmniAVS} provides 59,458 omnimodal expressions for 4,262 objects in 2,098 auditory videos.
The expressions integrate 4 different modalities: \textit{text}, \textit{speech}, \textit{sound}, and \textit{image}, forming 8 different referring expressions: 
{1)} Text;
{2)} Speech;
{3)} Text with Sound;
{4)} Speech with Sound;
{5)} Text with Image;
{6)} Speech with Image;
{7)} Text with Sound and Image; and
{8)} Speech with Sound and Image.
Each expression includes either text or speech, as they provide essential {instructions} for the task and other modalities

\vspace{-1mm}
\subsubsection{3D Scene}
\noindent$\bullet$~\textbf{ScanRefer}~\cite{scanrefer} comprises 51,583 natural language queries for 11,046 objects across 800 ScanNet~\cite{scannet} scenes, with an average of 13.81 objects and 64.48 queries per scene.

\noindent$\bullet$~\textbf{Instruct3D}~\cite{segpoint} consists of 280 scenes from ScanNet++\cite{scannet++} and 2,565 expression–object pairs, designed for 3D instruction-based (or reasoning-based) segmentation, with a focus on interpreting implicit user intent from natural language. 

\noindent$\bullet$~\textbf{Ref-LERF}~\cite{Ref-LERF} consists of 772 training images and 22 testing images sourced from LERF \cite{LERF}, and is newly annotated with 295 language expressions for 59 objects. The expressions focus on spatial relationships, providing rich contextual cues for referring segmentation in 3D Gaussian Splatting.

\vspace{-1mm}
\subsubsection{GREx Datasets.}
\noindent$\bullet$~\textbf{gRefCOCO}~\cite{gres} contains 278,232 referring expressions across 19,994 images, supporting both GRES~\cite{gres} and GREC~\cite{grec} tasks. It extends RefCOCO by adding multi-target and no-target expressions alongside single-target ones. No-target expressions are carefully curated to be contextually relevant yet intentionally misleading, without being entirely unrelated to the image. These characteristics make gRefCOCO more challenging than traditional RES datasets and more representative of real-world scenarios.

\noindent$\bullet$~\textbf{Multi3DRes}~\cite{3d-gres} extends Multi3DRefer~\cite{multi3drefer} by incorporating segmentation masks to support the 3D-GRES task~\cite{3d-gres}. It provides 61,926 expressions for 11,609 objects across 800 scenes, enabling more robust handling of real-world scenarios.

%% file: Secs/3_Meta_Architecture.tex
\section{Meta Architecture}
\label{sec:meta_architecture}

\subsection{Paradigm}
\label{sec3.1}
\noindent$\bullet$~\textbf{Two-Stage Paradigm.} Two-stage referring segmentation methods~\cite{mattnet,tgnn,cmn,cm-att,citd} follows a sequential process: the model first generates mask or tracklet proposals covering all objects in the scene or across the video, then matches these proposals to the referring expression, and finally selects the best-matched mask as the final prediction. These region proposals are typically obtained using off-the-shelf, well-trained instance segmentation models \cite{maskrcnn}. These candidate masks are evaluated based on their feature similarity to the referring expression. These two-stage methods, \eg, TGNN~\cite{tgnn} and MAttNet~\cite{mattnet}, have been widely adopted in early referring segmentation methods due to their modularity and interpretability. However, this paradigm is susceptible to error propagation, where inaccuracies in the proposal stage directly affect final performance. Moreover, it often incurs higher computational costs, limiting its practicality for real-time or resource-constrained applications.

\noindent$\bullet$~\textbf{One-Stage Paradigm.} One-stage methods~\cite{vlt,primitivenet,oneref,segllm,mcn,step} addresses the limitations of the two-stage methods by directly predicting the target object from the input visual scene and referring expression in a single forward pass. This end-to-end architecture eliminates the need for separate proposal generation and matching steps, potentially reducing error propagation and improving efficiency. One-stage methods typically employ dense prediction mechanisms where each spatial location in the feature map interacts with the referring expression. Following DETR~\cite{detr}, recent advances in transformer-based architectures, \eg, VLT~\cite{vlt,vlticcv} and ReLA~\cite{gres}, have significantly enhanced the performance by enabling more effective multimodal fusion and context modeling. These models can better capture the complex relationships between visual elements and linguistic descriptions, leading to more accurate segmentation results across various visual scenes.

\vspace{-1mm}
\subsection{Feature Extraction}
\label{sec3.2}
\noindent$\bullet$~\textbf{Vision Encoder.} Vision encoder extracts visual features from visual inputs. For image-based tasks, early methods~\cite{lstm-cnn,mattnet} rely on convolutional neural networks (CNN) such as ResNet~\cite{resnet}. Recently, Vision Transformers (ViT)~\cite{vit} and their variants~\cite{swin} have demonstrated superior performance and have been widely adopted by many methods~\cite{vlt,gres}. For video-based tasks, models~\cite{A2D&J-HMDB,Refer-Youtube-VOS} incorporate temporal modeling capabilities via 3D CNN~\cite{c3d,i3d}, or transformer with temporal attention, \eg, Video Swin Transformer~\cite{video_swin_transformer}. For 3D tasks~\cite{segpoint,refmask3d}, specialized encoders, \eg, Sparse 3D U-Net~\cite{sparse_3d_unet} and PointNet~\cite{pointnet}, process point clouds or voxel to capture spatial geometry and relationships.

\noindent$\bullet$~\textbf{Text Encoder.} Text encoders transform referring expressions into text features. Early methods~\cite{lstm-cnn,dmn,rmi,kwa,rrn} use recurrent neural networks to model sequential dependencies in language. 
With the advancement of pre-trained language models, recent methods~\cite{lavt,polyformer,uniref++,restr} have adopted transformer-based architectures, \eg, BERT~\cite{bert} and RoBERTa~\cite{roberta}, to obtain richer and more contextualized representations. More recently, text encoders from vision-language models, \eg, CLIP~\cite{clip}, have gained popularity for producing text embeddings that align well with visual features in a shared semantic space, enhancing multimodal alignment in referring segmentation~\cite{cris,etris}.

\noindent$\bullet$~\textbf{Audio Encoder.} Audio encoders extract acoustic features for audio-based tasks. Raw audio is typically converted into spectrograms or mel-frequency cepstral coefficients, which are then processed by neural networks. CNN-based models \cite{vggish} treat spectrograms as images, while transformer-based models~\cite{wav2vec,ssast} capture temporal dependencies in audio. Pretrained encoders like VGGish~\cite{vggish} and wav2vec~\cite{wav2vec} have shown strong performance in generating robust audio features for downstream tasks including AVS~\cite{avsbench}, Ref-AVS~\cite{refavs}, and OmniAVS~\cite{OmniAVS}.

\subsection{Multimodal Interaction}
\label{sec3.3}
\noindent \textbf{\textit{A. Multimodal Fusion}:} the process of integrating features from different modalities to create a unified representation.

\noindent$\bullet$~\textbf{Concatenation-based Fusion.} This fusion strategy~\cite{rmi,lstm-cnn,dmn,rrn,step} combines multimodal features via simple concatenation, followed by convolutional layers or MLPs. For example, image and text features are concatenated to form a joint representation: $f_\text{fused}\!=\!\text{Conv}(\text{Concat}[f_\text{image}; f_\text{text}])$. While efficient, such simple fusion fails to capture complex inter-modal interactions, limiting its effectiveness in fine-grained cross-modal reasoning.

\noindent$\bullet$~\textbf{Attention-based Fusion.} To address the limitations of simple fusion, attention-based methods~\cite{brinet,vlt,lavt,efn} enable dynamic and context-aware interactions across modalities. They selectively emphasize relevant features from one modality conditioned on the other~\cite{m3att}, facilitating fine-grained alignment. For example, in image-text tasks, attention guides the model to focus on specific word tokens when processing visual regions, and vice versa. Representative strategies include visual attention~\cite{vlt,vlticcv}, symmetric co-attention~\cite{etris,risclip}, and multimodal transformers employing self- and cross-attention~\cite{oneref,shared-ris}. These methods enhance models' ability to capture complex inter-modal relationships and have shown superior performance over simple fusion.

\vspace{1.6mm}
\noindent \textbf{\textit{B. Multimodal Alignment}:} the process of establishing meaningful correspondences between elements of different modalities.

\noindent$\bullet$~\textbf{Contrastive Learning-based Alignment.} These methods~\cite{cris,risclip,cm-masksd} leverage contrastive objectives to align multimodal representations in a shared embedding space. Methods like CLIP~\cite{clip} and ALIGN~\cite{align} train on paired image–text data to maximize the similarity of matched pairs and minimize that of mismatched pairs. This results in a unified semantic space where semantically related concepts from different modalities are closely positioned, supporting cross-modal retrieval and understanding.

\noindent$\bullet$~\textbf{Self-supervised Alignment.} These methods~\cite{transavs,slvp,oneref,risclip} leverage naturally co-occurring multimodal data to learn alignment without explicit supervision. Techniques include 1) masked multimodal modeling~\cite{oneref} that learns to predict masked content in one modality according to another modality and 2) reconstruction-based methods~\cite{slvp,risclip} that encode information from one modality and decode it into another.

\subsection{Temporal Information Processing}
\label{sec3.4}
For video tasks, temporal information processing is essential for motion understanding~\cite{MeViS} and consistent object segmentation across frames, especially under complex scenes~\cite{MOSE,MOSEv2}.

\noindent$\bullet$~\textbf{3D Convolutional Networks.} 3D CNNs~\cite{cmpc-pami,cmpc,A2D&J-HMDB} extend 2D convolutions by adding a temporal dimension to learn spatiotemporal features from videos. They process multiple frames simultaneously, capturing motion patterns and temporal context. Methods like C3D~\cite{c3d} and I3D~\cite{i3d} treat videos as 3D volumes, applying convolutions across spatial and temporal dimensions to extract features modeling object movements and transformations.

\noindent$\bullet$~\textbf{Temporal Attention Mechanisms.} Attention-based methods \cite{cmsa,efcma,locater,lbdt} capture temporal dependencies by dynamically weighting the importance of different frames. Temporal attention allows models to focus on the frames that are most informative for object tracking and segmentation, enhancing the handling of long-range dependencies and diverse motion patterns.

\noindent$\bullet$~\textbf{Memory Networks.} Memory-based methods \cite{Refer-Youtube-VOS,findtrack,referdino,vrs-hq} explicitly store and update objects' features across frames, capturing clues about their appearance, location, and context. These memory modules enable models to retrieve relevant historical cues when processing new frames, enhancing temporal consistency. Such mechanisms are particularly beneficial in complex and long videos, where target objects may temporarily disappear, reappear, or undergo significant appearance variations~\cite{MeViS,MOSE}.

\noindent$\bullet$~\textbf{Optical Flow and Motion Estimation.} A number of temporal modeling methods \cite{slvp,zhao2022modeling,losh,Refer-DAVIS,A2D&J-HMDB} leverage explicit motion cues by incorporating optical flow estimation. Optical flow captures pixel-level correspondences between consecutive frames, providing detailed information about object motion. This motion guidance can be effectively combined with appearance features to enhance tracking and temporal coherence of segmentation masks.

\subsection{Segmentation Head}
\label{sec3.5}
\vspace{-.6mm}
Segmentation head plays a key role by converting features into final masks. Existing designs can be broadly categorized as follows.\\
\noindent$\bullet$~\textbf{CNN-based Segmentation Head.} Traditional segmentation heads~\cite{lstm-cnn,rmi,step,rrn,cmsa} typically use a series of convolutional layers followed by upsampling operations as in FCN~\cite{fcn} to restore the spatial resolution of feature maps. 
\\\noindent$\bullet$~\textbf{Transformer-based Segmentation Head.} Prominent architectures such as DETR \cite{detr} and Mask2Former \cite{mask2former} employ transformer decoders to generate object queries, which are subsequently mapped to segmentation masks. Transformer-based segmentation heads \cite{referring_transformer,refmask3d,gres,transrmot} are particularly effective at capturing global context and are well-suited for complex scenes.

\noindent$\bullet$~\textbf{Promptable Segmentation Head.} Recent advances have introduced flexible and generalizable segmentation heads \cite{vrs-hq,sa2va,samwise,dit-sam,evf-sam,SAMA} that respond to diverse types of prompts. Segment Anything Model (SAM) \cite{sam} exemplifies this design by supporting point, box, and text-based prompts without requiring task-specific fine-tuning, enabling generalization across a wide range of segmentation tasks. Building on this foundation, SAM2 \cite{sam2} extends the promptable framework to video segmentation.

\subsection{Training Objectives}
\label{sec3.6}

Herein we list the most commonly used training objectives.

\noindent$\bullet$~\textbf{Segmentation Objectives.} For segmentation tasks~\cite{avsbench,lstm-cnn,vlt,vlticcv,gres}, the primary training objectives typically include binary cross-entropy (BCE) loss and Dice loss. BCE measures pixel-wise classification error, while Dice loss directly optimizes the overlap between predicted and ground truth masks. These losses are often combined to improve boundary accuracy and region completeness.

\noindent$\bullet$~\textbf{Grounding Objectives.} To enhance grounding performance, several methods~\cite{mcn,polyformer} incorporate grounding objectives (\eg, L1, L2, IoU, and focal loss) to enforce accurate correspondence between visual regions and referring expressions. Set-based bipartite matching~\cite{detr,grec} is used to address permutation invariance.

\noindent$\bullet$~\textbf{Multimodal Alignment Objectives.} Beyond visual perception losses, many referring segmentation models~\cite{vlt,cris,dshmp,refmask3d,soc,sun2024unveiling} incorporate alignment objectives to bridge visual and linguistic modalities. Contrastive losses are commonly used to pull together matched visual-language pairs and push apart mismatched ones. Effectively combining these objectives is essential for building robust models capable of accurate segmentation across diverse scenes and referring modalities.

\noindent$\bullet$~\textbf{Multi-Task Learning Objectives.} Some methods adopt multi-task learning strategies that couple referring segmentation with auxiliary tasks such as referring comprehension or generation. For example, MCN \cite{mcn} jointly models RES and REC with consistency and suppression objectives to reduce task conflict. Chen~\etal~\cite{lang2seg} integrate referring generation to enforce caption-aware consistency. Liu~\etal~\cite{m3att} introduce iterative language-vision interaction and reconstruction to preserve linguistic cues. These strategies show that jointly optimizing related tasks can enhance RES performance and multimodal understanding.

%% file: Secs/4_Image_Scene.tex
\section{Referring Expression Segmentation}
\label{sec:res}
\subsection{Two-stage Methods}
\label{sec4.1}
\begin{figure}[t]
    \label{fig:RES}
    \centering
    \includegraphics[width=1\linewidth]{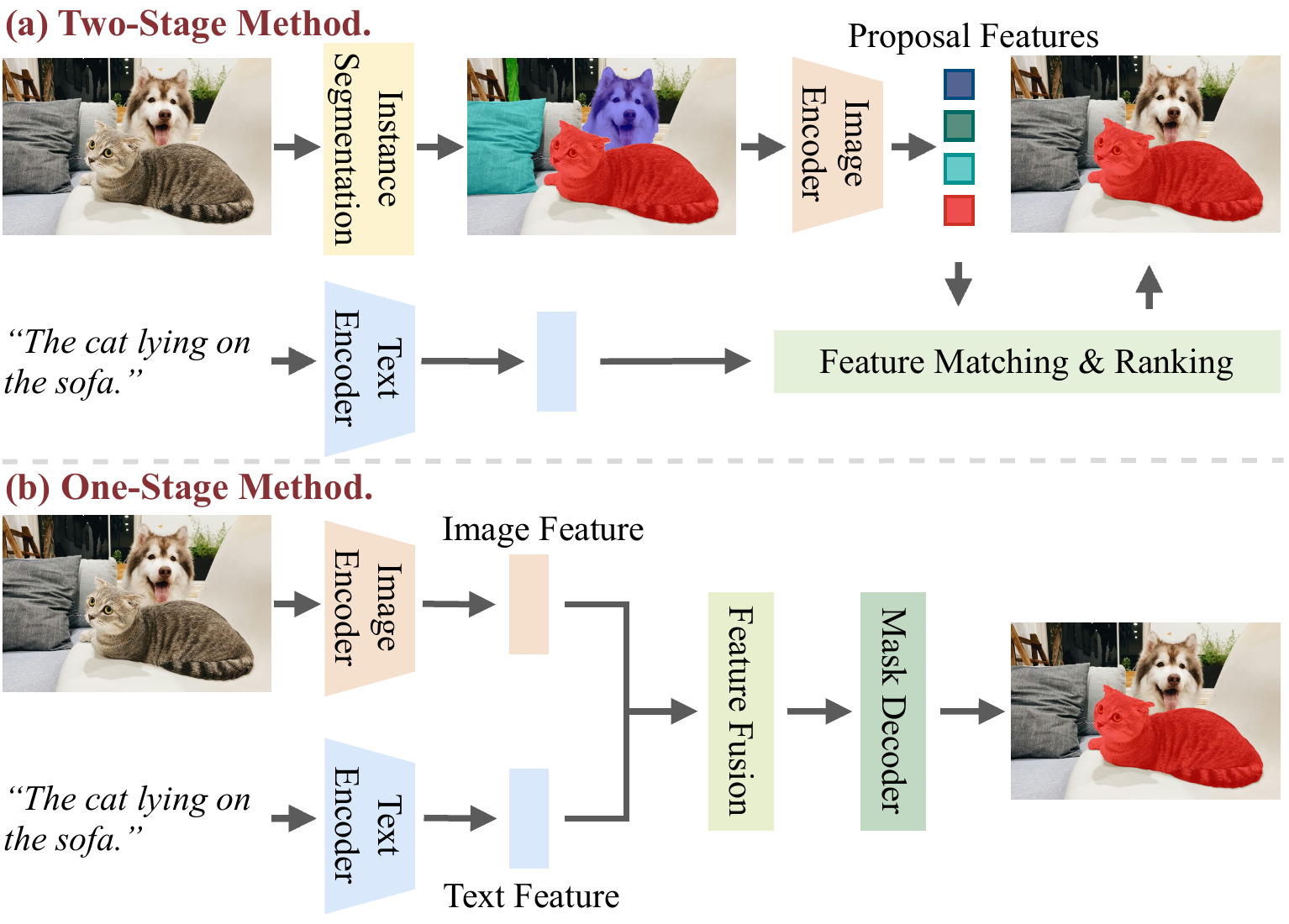}
    \vspace{-6mm}
    \caption{
        \textbf{Architecture Overview of Referring Expression Segmentation.} 
        (a) \textit{Two-stage method} uses an off-the-shelf instance segmentation model to generate region proposals, followed by vision-language feature matching and ranking to select the top-1 mask. 
        (b) \textit{One-stage method} fuses image and text features, performing pixel-level segmentation directly on the fused features. 
    }\label{Fig:RES_Architecture}
    \vspace{-3mm}
    
\end{figure}

Two-stage methods~\cite{mattnet,cmn,nmtree,RefCOCO&RefCOCO+&RefCOCO(g)} for RES typically involve an initial segmentation step followed by a matching process, as shown in \cref{Fig:RES_Architecture}(a). These methods often leverage off-the-shelf instance segmentation models to generate object proposals, and then match them to the referring expression to identify the best-matched object. 
For example, MAttNet~\cite{mattnet} first segments all objects using an instance segmentation network, Mask R-CNN~\cite{maskrcnn}, and then employs a modular network to identify the best-matched instance.
CMN~\cite{cmn} also adopts modular networks that parse referring expressions into subject, relationship, and object components using three soft attention maps. It then aligns these textual representations with image regions. 
ISF~\cite{isf} and WiCo~\cite{wico} combine the advantages of both two-stage and one-stage methods to improve the performance of RES.

However, this paradigm suffers from error accumulation and high computational cost, making it less practical in real-time settings. The majority of existing RES methods adopt a \textit{\textbf{one-stage}} paradigm, which we categorize in the following subsections.

\vspace{-1mm}
\subsection{One-stage Methods}
\label{sec4.2}
\subsubsection{Better Representations}
\noindent$\bullet$~\!\textbf{Optimizing Representation Extraction.}
Several works~\cite{tv-net,mcres,kwa} optimize feature extraction to obtain richer and more discriminative representations.
TV-Net~\cite{tv-net} enhances vision features of the referent by retrieving relevant external images.
KWA~\cite{kwa} enhances language feature by assigning higher weights to words critical for object identification. 
MCRES~\cite{mcres} builds virtual datasets and combines them with meta-learning optimization, enabling model to better capture the semantic and visual representations of concepts when handling expressions with novel compositions.
Some methods~\cite{asda,vg-law,vpd} focus on optimizing feature extraction to obtain language-guided vision representations or vision-guided language representations.
For example, ASDA~\cite{asda} adaptively selects the most relevant vision features according to sentence-level language cues.
VG-LAW~\cite{vg-law} treats the vision backbone as an expression-specific feature extractor by generating dynamic weights for various expressions.

\noindent$\bullet$~\!\textbf{Data Augmentation Methods.}
Data augmentation helps learn better representations by increasing the diversity and quantity of training data. However, standard techniques often distort spatial relationships or alter contextual cues described in referring expressions, making the augmented samples invalid. 
To address this, several works~\cite{nemo,maskris,clipseg,negationclip} propose RES-specific data augmentation techniques.
NeMo~\cite{nemo} constructs mosaic images by combining a target image with three CLIP-selected negative images, generating more diverse and challenging training examples. 
MaskRIS~\cite{maskris} introduces a masking-based data augmentation technique that combines both image and text masking to generate diverse image-text pairs for training.

\vspace{-1mm}
\subsubsection{Enhancing Multi-Modal Interaction}
\label{res-multimodal-interaction}

\noindent$\bullet$~\!\textbf{Multimodal Fusion.}
Various methods~\cite{lstm-cnn,vlt,s2rm} have been proposed to fuse vision and language features.
Early methods~\cite{lstm-cnn,rmi,rrn,dmn} typically extract vision features using CNNs and encode referring expressions via LSTMs. These features are then fused using simple operations such as concatenation, followed by convolutional layers. 
Kesen~\etal~\cite{kesen2022modulating} extend this by performing feature fusion in both the upsampling and downsampling paths of U-Net. 
With the advent of attention mechanisms, some works~\cite{m3att,cmsa,brinet,efn,bkinet} leverage cross-modal attention to capture fine-grained interactions between modalities. 
CMSA~\cite{cmsa} constructs comprehensive multimodal features by concatenating image, spatial, and language features, then processes them by a cross-modal self-attention module to model long-range dependencies between words and spatial regions.
EFN~\cite{efn} leverages an asymmetric co-attention module to enhance the matching between multi-modal features and strengthen their targeting capability.

A major milestone in RES is the shift toward Transformer-based architectures, which naturally support long-range dependencies across spatial and linguistic tokens. Traditional CNNs struggle with such global context due to their localized receptive fields. 
VLT~\cite{vlt,vlticcv} is the first to introduce a Transformer-based approach to RES, reformulating it as an attention problem where language features act as queries over vision tokens. 
This enables joint reasoning across both modalities. Since VLT, Transformer-based methods~\cite{lavt,slvit,lgformer,cgformer,lqmformer} have quickly become the common and state-of-the-art methods in RES. 
LAVT~\cite{lavt} injects language into vision features at intermediate levels of a vision Transformer to enhance multimodal fusion.
To mitigate query collapse, LQMFormer~\cite{lqmformer} enforces a margin of separation between query representations.
To address the high computational cost of Transformer, ReMamber~\cite{remamber} adopts a Mamba-based~\cite{mamba} architecture for more efficient vision-language fusion in RES.

\noindent$\bullet$~\!\textbf{Multimodal Alignment.}
Multimodal alignment~\cite{cris,risclip,cm-masksd,Segment_Anyword} in RES aims to explicitly or implicitly associate information from different modalities (\ie, language and vision), ensuring consistency within a shared semantic space. 
Several methods~\cite{risclip,caris,cm-masksd,restr,shared-ris} employ {cross-modal attention} to strengthen alignment across modalities. 
For example, ReSTR~\cite{restr} employs a self-attention encoder to effectively align visual and language features in a shared semantic space.

Other methods~\cite{cris,cgformer,magnet,coupalign,crossvlt} employ {contrastive learning} to improve multimodal alignment. By contrasting matched and mismatched modality pairs, these methods pull aligned pairs closer while pushing apart misaligned ones.
CRIS~\cite{cris} introduces a text-to-pixel contrastive loss to explicitly link textual and pixel-level visual features, addressing CLIP's limitations in RES. 
CGFormer~\cite{cgformer} integrates contrastive learning with a grouping strategy to associate tokens with corresponding masks. 
Some methods~\cite{btmae,oneref} adopt reconstruction-based techniques to align diverse modalities.
Inspired by MAE~\cite{mae}, BTMAE~\cite{btmae} introduces bidirectional reconstruction of missing features in both image and language tokens to effectively model high-dimensional relationships among multimodal tokens.
Inspired by BEiT-3~\cite{beit-v3}, OneRef~\cite{oneref} introduces a one-tower modality-shared Transformer with a mask referring strategy that jointly models referring-aware image and language masks.

\noindent$\bullet$~\!\textbf{Parameter-Efficient Tuning Methods.}
While large foundation models have advanced RES, fully fine-tuning remains computationally expensive. Parameter-Efficient Tuning (PET) offers a cost-effective alternative by freezing most of the pre-trained model and updating only a small subset of parameters, often achieving comparable performance. 
Recent works \cite{etris,barleria,detris,etrg} have explored applying PET to RES, mainly focusing on enhancing modal interaction while maintaining computational efficiency.
For example, ETRIS~\cite{etris} leverages a vision-language bridge to fuse visual inductive biases with linguistic cues,
while BarLeRIa~\cite{barleria} employs bi-directional intertwined vision-language adapters for efficient cross-modal fusion with minimal learnable parameters.

\vspace{-.9mm}
\subsubsection{Optimizing Mask Decoder}
To improve segmentation quality, several methods\cite{step,jmceln,mdsm,rrn,sadlr,convlstm,busnet} adopt multi-stage optimization strategies that progressively refine segmentation masks.
Early methods~\cite{rrn,step,lscm,cmpc,cmpc-pami} leverage ConvLSTM~\cite{convlstm} to iteratively decode multimodal features, refining the prediction mask.
SADLR~\cite{sadlr} iteratively refines predictions using accumulated object context to produce accurate segmentation masks.
JMCELN~\cite{jmceln} proposes a multi-stage cascade framework that refines segmentation by dynamically updating contextual embeddings based on intermediate mask predictions.

Segment Anything Model (SAM)~\cite{sam} shows strong performance across various segmentation tasks using point, box, or mask prompts, but remains limited in text-guided segmentation.
To bridge this gap, several methods~\cite{f-lmm,dit-sam,grounded_sam,prompt-ris,Segment_Anyword} have been proposed to adapt SAM for RES. 
DIT-SAM~\cite{dit-sam} addresses SAM’s limitations by projecting text into the semantic space of SAM’s image encoder.
Prompt-RIS~\cite{prompt-ris} bridges CLIP and SAM via prompt learning. 
Grounded SAM~\cite{grounded_sam} combines Grounding DINO~\cite{groundingdino} with SAM to enable detection and segmentation of arbitrary regions based on free-form text inputs.
F-LMM~\cite{f-lmm} uses a frozen LLM to generate segmentation priors, followed by a CNN-based and a SAM-based mask decoder to produce the final segmentation mask.
Recent works~\cite{lisa,gsva,sam4mllm} enhance SAM’s text-guided segmentation by generating prompt embeddings via LLMs (see \cref{res: rs}).

\vspace{-1.12mm}
\subsubsection{Improving Training Objectives}

Several methods~\cite{lts,pvd,addp,polyformer,cgan} focus on optimizing training objectives. LTS\cite{lts} decouples RES into a \textit{“Locate-Then-Segment”} framework, first predicting a position prior and then refining the segmentation mask. Other works~\cite{polyformer,partial-res,seqtr,pvd} reformulate segmentation as point-based sequence prediction, representing masks as polygons or point sequences. Text4Seg~\cite{text4seg} introduces a text-as-mask paradigm, casting RES as a text generation task without relying on additional decoders. UFO~\cite{ufo} treats segmentation as an embedding retrieval problem, generating masks by computing similarity between mask token embeddings and image features. Training objectives for enhancing multimodal interaction are discussed in Sec.~\ref{res-multimodal-interaction}.

\vspace{-.9mm}
\subsection{Multi-Task Learning}
\label{sec4.3}

Multi-task learning is widely used in segmentation, detection, and generation, typically with a shared backbone and task-specific heads. Building on this paradigm, several methods~\cite{mcn,WeakMCN,segvg,PLVL} jointly address RES and referring expression comprehension (REC).
MCN~\cite{mcn} implements an explicit constraint strategy by introducing consistency loss to ensure similarity between feature activation maps in REC and RES tasks. Referring Transformer~\cite{referring_transformer} and MDETR~\cite{mdetr} use an implicit approach by sharing multi-modal representations across REC and RES heads.
SeqTR~\cite{seqtr} and Polyformer~\cite{polyformer} further unify both tasks as sequence-to-sequence point prediction problems. Beyond REC and RES, some works~\cite{lang2seg,xxxxvilam} also explore joint modeling of RES and referring expression generation (REG).

Generalist models supporting multiple vision-language tasks \cite{x-decoder,seem,glee,psalm,UniVG,UNINEXT,Pixel-SAIL} have shown strong performance on RES. X-Decoder~\cite{x-decoder} introduces a unified framework for segmentation and vision-language tasks, including RES. SEEM~\cite{seem} extends this versatility by supporting diverse prompts (clicks, boxes, polygons, scribbles, text, and image regions) via a prompt encoder in a joint visual-semantic space. 
AnyRef~\cite{anyref} leverages MLLMs for pixel-level grounding and region-aware expression generation across text, regions, images, and audio.

\subsection{Weakly-Supervised Methods}
\label{sec4.4}
To reduce the annotation burden, weakly-supervised RES methods~\cite{tseg,ppt,sag,pks} leverage incomplete, imprecise, or noisy annotations to minimize reliance on dense pixel-level labels.

\noindent$\bullet$~\!\textbf{Text-Only Supervision.}
TSEG~\cite{tseg} is a pioneering weakly supervised RES method that only uses image-level referring expressions as supervision. 
Subsequent methods~\cite{weakly-ris,sag,GroupViT,DViN,gbs,ppt} adopt textual supervision and employ strategies such as visual entity discovery and gathering~\cite{sag}, text-to-image response mapping~\cite{tris}, and enhanced Grad-CAM for saliency refinement~\cite{weakly-ris}.
TRIS~\cite{tris} directly learns the text-to-image response map by contrasting target-related positive texts with target-unrelated negative texts.
PPT~\cite{ppt} leverages a point generator to connect frozen CLIP and SAM models while adopting curriculum learning to facilitate the gradual learning of the point generator.
PCNet~\cite{pcnet} decomposes the description into multiple cues to guide progressive target localization.
WeakMCN~\cite{WeakMCN} jointly learns WeakREC and WeakRES in a collaborative manner.
\\\noindent$\bullet$~\!\textbf{Bounding Box Supervision.} Feng~\etal~\cite{feng2022learning} propose a weakly supervised RES method using bounding box annotations, where pseudo labels are generated from object contours and refined through filtering. In contrast, GTMS~\cite{gtms} enhances pseudo label quality by incorporating both structural and semantic information.

\noindent$\bullet$~\!\textbf{Point Supervision.} 
Beyond text and bounding box, PKS~\cite{pks} employs click-based annotations (\ie, object center/corner clicks) for model training, achieving commendable performance.

\vspace{-1.16mm}
\subsection{Semi-Supervised Methods}
\label{sec4.5}
Semi-supervised RES methods reduce reliance on labor-intensive annotations by leveraging a small set of labeled image-text pairs alongside abundant unlabeled samples. 
Several semi-supervised RES methods~\cite{semires,pseudo-ris} adopt pseudo-labeling strategies, selecting high-confidence predictions as supervision for model refinement. 
RESMatch~\cite{resmatch} introduces the first dedicated semi-supervised RES framework, incorporating a quality assessment mechanism to evaluate pseudo-labels and strong–weak supervision pairs. 
SemiRES~\cite{semires} leverages SAM’s edge-segmentation capabilities to generate high-quality pseudo-labels. 
Pseudo-RIS~\cite{pseudo-ris} produces multiple candidate masks from distinctive words and filters misleading captions to obtain reliable supervisory signals.
Some semi-supervised RES methods~\cite{partial-res,safari} adopt a mixed-supervision paradigm, typically using a small portion of mask annotations alongside a larger proportion of bounding box annotations as supervisory signals.
Partial-RES~\cite{partial-res} and Safari~\cite{safari} adopt an auto-regressive contour-based sequence prediction strategy, requiring only a small fraction of mask annotations along with supplementary bounding box annotations.

\subsection{Zero-Shot Methods}
\label{sec4.6}
Zero-shot RES methods~\cite{global-local-clip,iterprime,peekaboo,RESAnything} leverage vision-language foundation models (\eg, CLIP~\cite{clip}) to perform segmentation without task-specific training.
Global-Local CLIP~\cite{global-local-clip} is the first method to explore zero-shot RES using CLIP.
TAS~\cite{tas} incorporates additional caption embeddings, negative text embeddings, and a spatial rectifier to enhance CLIP predictions, while CaR~\cite{CaR} recurrently applies CLIP to iteratively refine the segmentation mask.
HybridGL~\cite{HybridGL} introduces a global-local feature extraction method that combines mask-specific details with contextual information to improve mask representation.

\vspace{-1.6mm}
\subsection{Other Task Settings}
\label{sec4.7}
\label{res: rs}

\noindent$\bullet$~\!\textbf{Reasoning Segmentation.} 
Large language models (LLMs) and multimodal LLMs (MLLMs) \cite{gpt4,llava,internvl} have significantly advanced vision-language tasks by enabling strong commonsense reasoning, opening new opportunities for RES.
Building on this, LISA~\cite{lisa} pioneers the concept of reasoning segmentation, allowing models to handle complex expressions requiring external knowledge, such as ``\textit{Segment the food with the highest protein content}.'' 
LISA introduces an embedding-as-mask paradigm by extending the MLLM’s vocabulary with a special [\texttt{SEG}] token to prompt SAM~\cite{sam} for mask generation.
Inspired by these developments, several methods~\cite{reasoningseg_survey,pixellm,RSVP,PRIMA,pixfoundation,osprey,InstructSeg,next-chat,PerceptionGPT} explore reasoning segmentation with LLMs/MLLMs. 
CoReS~\cite{CoReS} employs chain-of-thought (CoT)~\cite{cot} to tackle complex implicit text queries that require multi-step reasoning.
SAM4MLLM~\cite{sam4mllm} encodes object masks as discrete text prompts, while READ~\cite{READ} converts text-image similarity maps into differentiable points to prompt SAM.
Unlike LISA's paradigm, LLM-Seg~\cite{llm-seg} adopts a two-stage approach, decoupling the reasoning over text from the segmentation process.

Several methods~\cite{gsva,lasagna,SESAME,MMR} focus on addressing the challenges of multi-target or zero-target scenarios in images, known as GRES~\cite{gres}.
GSVA~\cite{gsva} addresses the GRES~\cite{gres} task by introducing shared-weight [\texttt{SEG}] tokens for multi-target segmentation and a [\texttt{REJ}] token to discard empty targets. 
SESAME~\cite{SESAME} handles false-premise queries by enabling models to detect object presence, provide corrective feedback, and segment only when appropriate.
In parallel, interactive and conversational segmentation has gained interest~\cite{segllm,MIRAS}.
For example, SegLLM~\cite{segllm} supports multi-turn interactions with visual and textual queries, enabling the inference of object relationships such as spatial, interactive, and hierarchical dependencies.
More recently, methods~\cite{Seg-Zero,POPEN,PixelThink,SAM-R1} have explored reinforcement learning (RL) to enhance reasoning capabilities in segmentation. Seg-Zero\cite{Seg-Zero} employs pure RL to develop emergent reasoning and improve out-of-domain generalization. POPEN\cite{POPEN} incorporates preference-based optimization to align large vision-language models with human intent via RL strategies, such as GRPO~\cite{deepseekmath}.

\noindent$\bullet$~\!\textbf{Referring Remote Sensing Image Segmentation (RRSIS).} 
RRSIS focuses on aerial or satellite images, posing unique challenges by varying spatial scales, object orientations, and complex backgrounds distinct from natural images. 
To address these issues, several tailored methods~\cite{rrsis,danet,rmsin,geochat,vrsbench} are proposed. 
Yuan~\etal~\cite{rrsis} introduce RRSIS task and RefSegRS benchmark, along with a LAVT-based~\cite{lavt} framework that integrates multi-scale features for segmenting small and scattered objects. 
RMSIN~\cite{rmsin} addresses orientation variations using rotated convolutions to better capture multi-oriented object features. 

\noindent$\bullet$~\!\textbf{Other.} 
Beyond the core tasks, alternative settings have been explored \cite{rewatbowornwong2023zero,glamm,cvmn,OMGLLaVA}. 
Zero-guidance segmentation \cite{rewatbowornwong2023zero} seeks to segment input images and label all segments using natural language.
Grounded Conversation Generation \cite{glamm} aims to generate natural language responses interleaved with object segmentation masks.
Several methods~\cite{clipu2net,clipunetr} extend referring segmentation capabilities to embodied intelligence settings.
Additionally, PTQ4RIS~\cite{ptq4ris} proposes a post-training quantization framework to address challenges for on-device RES inference.

%% file: Secs/5_Video_Scene.tex
\section{Referring Video Object Segmentation}
\label{sec:rvos}

\subsection{Per-frame and Online Methods}
\label{sec5.1}

\noindent$\bullet$~\!\textbf{Per-frame Methods.}
A natural approach to Referring Video Object Segmentation (RVOS) is to treat a video as a sequence of images and apply image-level referring segmentation to each frame independently. RefVOS \cite{refvos} follows this paradigm by fusing frame-level visual and language features using an image-based RES network. Subsequent methods such as CMPC~\cite{cmpc-pami} and VLT~\cite{vlt} adopt similar per-frame pipelines, offering a simple extension of image-level RES to the video domain.

\noindent$\bullet$~\!\textbf{Online Temporal Understanding.} Per-frame methods process each frame independently and often suffer from temporal inconsistency due to the lack of temporal context. To address this, online methods process videos sequentially while maintaining memory across frames. For example, URVOS~\cite{Refer-Youtube-VOS} uses a temporal memory attention module to improve intra-frame consistency, and OnlineRefer~\cite{onlinerefer} introduces an online association framework to enhance temporal coherence and referring accuracy.

\vspace{-1.16mm}
\subsection{Offline One-stage Methods}
\label{sec5.2}
\noindent$\bullet$~\!\textbf{Offline Temporal Understanding.}
Although online methods leverage historical frame information, their inability to access future frames restricts their capacity to resolve motion-centric and temporally ambiguous expressions, \eg, “\textit{Elephant that goes towards then turns back}”. To mitigate this limitation, many RVOS approaches employ offline processing, which enables global temporal reasoning by considering the full video sequence \cite{tf2,htr,tempcd,soc,cmsa,onlinerefer,locater,lbdt,lastc,html,vlp-rvos,findtrack,wang2020context,hui2021collaborative}. For example, TempCD~\cite{tempcd} employs global referent tokens and local object queries to perform video-level reasoning and frame-wise segmentation via a collection-distribution mechanism. HTML~\cite{html} applies hierarchical temporal sampling to capture multi-scale temporal interactions. LBDT~\cite{lbdt} performs early-stage spatio-temporal alignment using language-guided encoding, while LOCATER~\cite{locater} utilizes a finite-memory structure to dynamically gather relevant temporal context. ReferMo~\cite{Long-RVOS} employ motion-vectors to compress the motion in across the video. Most recent RVOS works follow this offline pipeline, with various designs optimizing different aspects of video-level reasoning.

\noindent$\bullet$~\!\textbf{Better Representations.} Some RVOS methods~\cite{mcintosh2020visual,mlrl,manet,losh,vd-it,dshmp,acga} focus on learning better representations to enhance model performance. They improve either visual or textual feature representations to strengthen multi-modal understanding. On the visual side, several methods~\cite{mcintosh2020visual,mlrl,manet,vd-it,acga} aim to enhance visual feature quality. MLRL~\cite{mlrl} introduces a multi-level representation learning framework that generates discriminative embeddings by integrating multi-frame temporal dynamics (video level), spatial context (frame level), and object-aware priors (object level). VD-IT~\cite{vd-it} leverages visual features from pretrained text-to-video diffusion models, achieving improved temporal consistency across frames. On the textual side, methods such as LoSh~\cite{losh} and DsHmp~\cite{dshmp} focus on refining language representations. LoSh~\cite{losh} jointly predicts with long and short expressions, using the short form to enhance appearance-based localization. To address the challenging motion expressions~\cite{MeViS}, DsHmp~\cite{dshmp} decomposes referring expression understanding into static and motion perception components, with an emphasis on improving temporal language comprehension. Some other recent works explore using extra types of feature, such as spectrum~\cite{sgmg} or flow map~\cite{zhao2022modeling}, to assist in representation.

\noindent$\bullet$~\!\textbf{Enhancing Multi-Modal Interaction.}
Effective integration of visual and textual features is essential for RVOS. Several methods \cite{yofo,efcma,cmsa,oatnet,referformer,dmformer} have been proposed to enhance cross-modal interaction. OATNet~\cite{oatnet} concatenates visual and textual features and processes them through a shared multi-modal encoder to jointly model intra- and inter-modal relationships. EFCMA~\cite{efcma} introduces an encoder fusion network with gradual language guidance and a co-attention mechanism to enhance feature alignment. ReferFormer~\cite{referformer} designs a cross-modal feature pyramid network to enable multi-scale fusion of vision-language features.
CMSA \cite{cmsa} introduces a gated multi-level fusion module to selectively integrate cross-modal features across visual hierarchies. YOFO \cite{yofo} designs a meta-transfer module that injects target-specific linguistic cues into visual features while suppressing irrelevant language variations. SSA~\cite{SSA} addresses the gap between linguistic descriptions and video objects, as well as interference from background clutter.

\noindent$\bullet$~\!\textbf{Optimizing Mask Decoder.}
Vision foundation models (\eg, SAM \cite{sam}, SAM2~\cite{sam2}) have shown strong segmentation capabilities, inspiring their adaptation for RVOS. Recent methods \cite{samwise,refsam,referdino,mpg-sam_2} leverage the precise mask generation of these models while extending them to handle referring expressions in video contexts. RefSAM~\cite{refsam} integrates multi-view information from diverse modalities and frames across time to adapt SAM for RVOS. SAMWISE~\cite{samwise} and MPG-SAM 2~\cite{mpg-sam_2} are two representative RVOS works built on SAM2~\cite{sam2}. SAMWISE~\cite{samwise} injects temporal cues and multimodal signals via an adapter during feature extraction, while MPG-SAM 2~\cite{mpg-sam_2} employs mask prior-based dense prompts and multi-level global context fusion.

\noindent$\bullet$~\!\textbf{Improving Training Objectives.}
Some methods~\cite{soc,slvp} aim to improve training objectives in RVOS. For example, SOC \cite{soc} introduces a visual-linguistic contrastive loss that applies semantic supervision at the video level to align object representations across modalities. Mei~\etal~\cite{slvp} propose a general self-supervised language-video pretraining framework that learns pixel-level features by using optical flow as an intermediate supervision signal during pretraining. Some methods~\cite{sa2va,univs,unimm,mutr,OMGSeg,uniref++,gpt-ps} address multiple video segmentation tasks via unified frameworks or multi-task learning strategies. MUTR~\cite{mutr} adopts a DETR-style transformer to jointly handle various tasks, while AL-Ref-SAM 2~\cite{gpt-ps} explores a training-free paradigm to unify RVOS and AVS. UniMM~\cite{unimm} presents a unified model for both mask-referred (VOS) and language-referred (RVOS) segmentation. UniVS~\cite{univs} unifies all major video segmentation tasks within a single model. Sa2VA~\cite{sa2va} integrates SAM2 with LLaVA to enable dense grounded understanding across text, image, and video in a shared LLM token space. Moreover, recent study such as VEGGIE~\cite{VEGGIE} shows that generalist generative models can also be used for RVOS.

\vspace{-1.16mm}
\subsection{Two-Stage Methods}
\label{sec5.3}

Similar to RES, many early RVOS methods adopt a two-stage paradigm. Some directly extend image-based methods, for example, Khoreva \etal~\cite{Refer-DAVIS} adapt image method MAttNet~\cite{mattnet} to videos by applying frame-wise segmentation followed by post-processing temporal smoothing. Others~\cite{citd, mttr} operate at the video level by generating object tracklets across the entire video sequence and selecting the one that best matches the referring expression. For example, Liang \etal~\cite{citd} first detect tracklets and then perform language-tracklet matching to identify the target.

\noindent$\bullet$~\!\textbf{Resurgence for Long and Motion Expressions.} Two-stage methods were initially in the minority due to the architectural complexity introduced by decoupling the segmentation and language-vision understanding stages. While this separation increases design complexity, it offers a key advantage: the generated tracklets can encode the entire temporal sequence in a compressed form, preserving both short-term and long-term motion cues for downstream multi-modal understanding in the second stage.

The importance of such global temporal modeling is highlighted by the challenging MeViS \cite{MeViS} benchmark, which emphasizes motion-centric expressions. Many expressions in MeViS describe long-term motions, requiring models to capture extended temporal dynamics. However, online methods lack access to global context, and offline one-stage models often rely on sparse frame sampling (\eg, 5 or 8 frames) to reduce computational cost, which can miss fine-grained or long-range motion patterns, leading to suboptimal performance. This challenge has motivated a resurgence of two-stage designs. Recent works~\cite{sola,DMVS} first extract full-length mask tracklets using off-the-shelf video instance segmentation models, then align them with language in a separate stage. This enables comprehensive temporal modeling and more accurate understanding of motion-guided expressions.

\vspace{-1.16mm}
\subsection{Other Task Settings}
\label{sec5.4}
\noindent$\bullet$~\!\textbf{Weakly-Supervised RVOS.}
RVOS methods typically depend on densely annotated datasets like MeViS~\cite{MeViS}, which are expensive to create. To reduce annotation costs, weakly supervised methods~\cite{simrvo,groprompt} have emerged. For example, SimRVO~\cite{simrvo} introduces a weak supervision scheme where only the first frame has mask supervision, and subsequent frames use bounding boxes.

\noindent$\bullet$~\!\textbf{Video Reasoning Segmentation.}~
This~setting~focuses~on~implicit expressions that require complex reasoning, \eg, “\textit{winner in the sprint race},” posing significant challenges in intent understanding and external knowledge. Driven by advances in LLMs/MLLMs~\cite{llava,gpt4,internvl}, several methods~\cite{visa,villa,videolisa,trackgpt,vrs-hq,InstructSeg} have been proposed to address this setting. They typically input video frames and queries into LLMs/MLLMs to generate \texttt{<SEG>} tokens, which are then decoded into segmentation masks. VideoLISA~\cite{videolisa} adopts a sparse-dense sampling strategy to balance temporal context and spatial detail under limited computation. VRS-HQ~\cite{vrs-hq} fuses spatial features into temporal tokens and employs SAM2~\cite{sam2} for keyframe segmentation followed by mask propagation. JiT~\cite{JiT} proposes an online method using ``digital twins'' to help LLM better understand video context. GLUS~\cite{GLUS} propose to migrate local and global reasoning with one transformer network.

\noindent$\bullet$~\!\textbf{Other Variants and Settings.}
Recent studies explore alternative RVOS variants \cite{SAMA,videoglamm,stbridge,fs-rvos,actionvos}. SAMA~\cite{SAMA}, MoRA~\cite{MoRA}, and ViCaS~\cite{ViCaS} extend the visual questioning and answering to segmentation answers, enables multi-turn dialogue or fine-grained segmentation. VideoGLaMM~\cite{videoglamm} leverages LLMs for grounded conversations, requiring text responses anchored at the pixel level in video frames. FS-RVOS~\cite{fs-rvos,FS-RVMOS} introduces a few-shot setting~\cite{MOVE} to address scenarios with limited annotated samples. ActionVOS~\cite{actionvos} focuses on egocentric videos, using action narrations as additional language prompts to segment active objects. Beyond general-scene video, RVOS can also be adapted to specific videos, such as medical or sugical videos~\cite{ReSurgSAM2,TPP}.

\vspace{-1.16mm}
\section{Referring Audio-Visual Segmentation}
\label{sec:r-avs}
\subsection{Audio-Visual Segmentation}
\label{sec6.1}
\subsubsection{Fully Supervised Learning}
\noindent$\bullet$~\!\textbf{Better Representation.}
Feature representations extracted from audio and visual encoders are critical to AVS. Recent works~\cite{teso,combo,qdformer,catr,lavish,selm} focus on enhancing these representations. Lin~\etal~\cite{lavish} demonstrate that Vision Transformers serve as parameter-efficient audio-visual learners. TeSO~\cite{teso} enhances audio guidance by generating scene-level descriptions with LLMs and extracting sounding-object cues via Chain-of-Thought prompting. QDFormer~\cite{qdformer} uses product quantization to disentangle multi-source semantics into noise-suppressed components and applies global-to-local distillation to refine frame-level audio features.
COMBO~\cite{combo} leverages foundation model priors for precise representations, while ECMVAE~\cite{ecmvae} factorizes features into shared and modality-specific components.

\noindent$\bullet$~\!\textbf{Enhancing Multi-Modal Interaction.}
Recent methods~\cite{deepavfusion,gavs,c3n,combo,catr,avsegformer} focus on improving audio-visual fusion and alignment. DeepAVFusion~\cite{deepavfusion} performs early fusion using learnable tokens to integrate audio-visual patches in parallel or sequentially.
GAVS~\cite{gavs} adopts an encoder–prompt–decoder framework to leverage SAM's generalization capacity. Dolphin~\cite{Dolphin} achieves fine-grained spatial-temporal alignment via multi-scale adapters and interleaved fusion, projecting features into an LLM for joint understanding. RAVS \cite{RAVS} mitigates audio ambiguity by clustering visual features by semantic density, weighting them by audio responsiveness, and modeling uncertainty for rapid transitions. DDESeg~\cite{DDESeg} disentangles mixed audio into semantic cues, filters noise via visual context, and enhances alignment through modality-specific discriminability.

\noindent$\bullet$~\!\textbf{Temporal Processing.}
Several AVS methods~\cite{avsac,avs-bigen,ufe,combo,avs-mamba} explicitly address temporal modeling. COMBO~\cite{combo} enforces temporal coherence via an adaptive inter-frame consistency loss based on cosine similarity of adjacent masks. UFE \cite{ufe} employs temporal partitioning strategy, using neighboring frames for motion guidance and distant ones to enhance data diversity.

\noindent$\bullet$~\!\textbf{Improving Training Objectives.}
Some methods~\cite{cpm,transavs,pif,avsc,bavs,avs-bigen,sun2024unveiling} aim to improve AVS performance by optimizing training objectives. PIF~\cite{pif} decomposes AVS into two subtasks: correlation learning, which aligns audio with visible individuals to provide positional priors, and segmentation refinement, which produces masks. While transformer-based methods~\cite{avsegformer} have advanced AVS, they suffer from cross-attention inefficiency and unstable bipartite matching. CPM~\cite{cpm} addresses these limitations by introducing a hybrid query strategy combining class-agnostic and class-conditional queries, and further enhances training with a prompt-based joint audio-visual contrastive objective. In contrast, TransAVS~\cite{transavs} improves audio query diversity using self-supervised losses at both query and mask levels. Other methods~\cite{bavs,avsc} introduce silent-object-aware objectives to ensure the segmentation of all potential sounding objects.

\vspace{-1mm}
\subsubsection{Weakly Supervised/Unsupervised Methods}

\noindent$\bullet$~\!\textbf{Weakly-Supervised AVS.} WS-AVS \cite{ws-avs} proposes a weakly-supervised method using instance-level annotations and a multi-scale contrastive learning to improve cross-modal alignment.

\noindent$\bullet$~\!\textbf{Unsupervised AVS.} Recent woks such as MoCA \cite{moca} leverage foundation models (\eg, SAM \cite{sam}, ImageBind \cite{imagebind}) to achieve competitive AVS performance without manual annotations.

\subsection{Referring Audio-Visual Segmentation}
\label{sec6.2}
Ref-AVS \cite{refavs} extends referring segmentation to the audio-visual domain, aiming to segment specific objects in video using synchronized audio and free-form language expressions. To support this task, Wang \etal~\cite{refavs} introduce the first Ref-AVS benchmark with over 3,000 videos annotated with masks and text expressions. The proposed framework employs modality-specific encoders for audio, vision, and text, fused via a query-based decoder. Ref-AVS handles scenarios beyond the scope of traditional AVS or RVOS, such as “\textit{person singing bass in a cappella group}”, by leveraging multimodal cues for fine-grained disambiguation. It also shows that adding language improves the discriminative power of audio cues~\cite{teso}. Omni-R1~\cite{Omni-R1} further enhances performance on Ref-AVS using reinforcement learning to select keyframes and rewrite tasks, enabling efficient Ref-AVS with just one training epoch. STBridge~\cite{stbridge} bridges the modality gap between speech and text, allowing RVOS models to effectively process noisy spoken input.
SAM2-LOVE~\cite{SAM2-LOVE} introduces a multimodal fusion framework that integrates audio, text, and visual information into learnable tokens to prompt SAM2~\cite{sam2} for Ref-AVS.

Despite its potential, Ref-AVS is still in its early stages. Challenges include temporal misalignment, underused spatial audio, and limited dataset diversity. Future work may explore spatialized audio, adaptive fusion, and broader benchmarks.

\begin{figure}[t]
	\centering
	\includegraphics[width=1\linewidth]{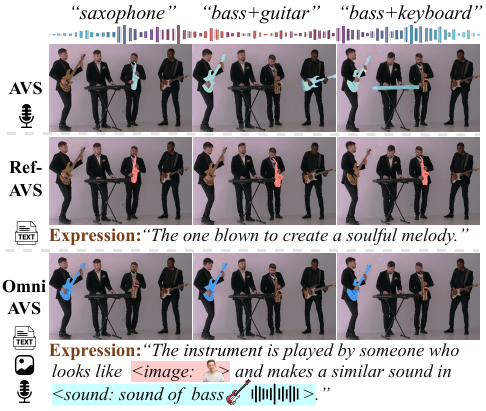}
        \vspace{-6mm}
	\caption{{Comparison of different audio-visual segmentation tasks. }}
    \vspace{-1\baselineskip}
    \label{fig:RAVS}
\end{figure}

\vspace{-1.6mm}
\subsection{Omnimodal Referring Audio-Visual Segmentation}
\label{sec6.3}

Omnimodal Audio-Visual Segmentation (OmniAVS) \cite{OmniAVS} is a recently proposed task that extends traditional text-only referring audio-visual segmentation by introducing omnimodal expressions, flexibly combining text, speech, sound, and visual cues, see \cref{fig:RAVS}. This formulation yields 8 types of expressions: 1) Text, 2) Speech, 3) Text + Sound, 4) Speech + Sound, 5) Text + Image, 6) Speech + Image, 7) Text + Sound + Image, and 8) Speech + Sound + Image. To support this task, Ying \etal~\cite{OmniAVS} introduce the first benchmark, OmniAVS, comprising 59,458 omnimodal expressions for 4,262 objects across 2,098 auditory videos. They propose Omnimodal Instructed Segmentation Assistant (OISA) as baseline, integrating a MLLM with a flexible mask head. OISA uses an audio-visual interleaving strategy to align content across modalities and generates target-aware [\texttt{seg}] tokens for segmentation, enabling accurate tracking and understanding in complex scenes.

While OISA demonstrates promising performance, several open challenges remain in OmniAVS:
\textbf{(i)} Improving audio-visual fusion via more robust joint representations, rather than relying on post-hoc alignment;
\textbf{(ii)} Disentangling overlapping sound sources when multiple objects emit sounds simultaneously;
\textbf{(iii)} Effectively combining multiple modalities within expressions through unified representations or cross-modal fusion;
\textbf{(iv)} Enhancing segmentation robustness in complex scenarios such as occlusion, disappearance, and reappearance;
\textbf{(v)} Pretraining on larger-scale audio-visual datasets to improve generalization to unseen settings.

%% file: Secs/6_3D_Scene.tex
\vspace{-1.16mm}
\section{Referring 3D Segmentation}
\label{sec:3d-res}

\subsection{3D Referring Expression Segmentation}
\label{sec7.1}

\subsubsection{Fully Supervised Learning} 

\noindent$\bullet$~\textbf{Two-Stage Methods.} Two-stage methods first generate object proposals via 3D instance segmentation, followed by language-guided identification of the target object. TGNN~\cite{tgnn} is a pioneering method in this paradigm, extracting high-quality instance masks and features, then modeling instance–language relationships using a Graph Neural Network, followed by multimodal feature aggregation for final prediction. Following this, X-RefSeg3D \cite{x-refseg3d} enhances cross-modal interaction by constructing a scene graph that integrates linguistic entities into visual representations, and employs dual-stream relation reasoning through textual and spatial modules to more effectively identify the target object.

\noindent$\bullet$~\textbf{One-Stage Methods.} One-stage methods perform end-to-end segmentation by directly fusing visual and linguistic features, offering improved efficiency over two-stage pipelines. 3D-STMN~\cite{3d-stmn} introduces a superpoint-text matching (STM) mechanism to align superpoints with referring expressions, enabling faster inference and improved multimodal alignment. LESS~\cite{less}, built on SPFormer~\cite{SPFormer}, adopts a one-stage pipeline and introduces an area regularization loss and point-to-point contrastive loss to mitigate interference from surrounding objects and background clutter. RefMask3D~\cite{refmask3d} proposes a Linguistic Primitives Construction (LPC) module to learn fine-grained semantic primitives, enhancing vision-language alignment during decoding. RG-SAN~\cite{Rg-san} improves spatial reasoning by integrating text-driven localization with rule-guided weak supervision to model inter-object spatial relationships. IPDN~\cite{ipdn} incorporates multi-view image features to enrich 3D representations and employs task-guided prompts to focus on language-relevant targets, addressing point cloud incompleteness and semantic ambiguity. While most methods focus on single-object segmentation, 3D-GRES~\cite{3d-gres} extends GRES~\cite{gres} to support an arbitrary number of target objects in point clouds, enabling generalized 3D referring segmentation.

\noindent$\bullet$~\textbf{3D Reasoning Segmentation.} Motivated by the success of LLMs/MLLMs~\cite{llava,internvl} in 2D reasoning segmentation~\cite{lisa}, recent studies have begun exploring their potential in 3D domain. SegPoint~\cite{segpoint} and Reason3D~\cite{reason3d} are early efforts that integrate LLMs/MLLMs to enable reasoning in 3D-RES, primarily focusing on single-object or single-category settings. To handle more complex scenarios, MORE3D~\cite{MORE3D} extends this capability to multi-object reasoning and additionally generates textual explanations alongside segmentation outputs. 3D-LLaVA~\cite{3D-LLaVA} further introduces a unified vision-language framework for 3D tasks, jointly addressing question answering, dense captioning, and referring segmentation while achieving competitive performance.

\noindent$\bullet$~\textbf{Multi-Task Learning.} Joint learning of 3D Referring Segmentation (3D-RES) and Referring Expression Comprehension (3D-REC) is a natural extension of multi-task learning, as demonstrated in 2D vision~\cite{mcn,seqtr}, where shared representations benefit related tasks. 3DRefTR~\cite{3dreftr} follows this paradigm by extending a 3D-REC model through reuse of query embeddings and visual tokens, adding a mask prediction branch to support 3D-RES with minimal overhead. MCLN~\cite{mcln} further explores task synergy by jointly optimizing separate branches for 3D-REC and 3D-RES, leveraging their semantic and structural alignment.

Beyond pairwise integration, recent efforts aim to develop unified models for a broad range of 3D vision-language tasks. Uni3DL~\cite{uni3dl} introduces a versatile framework that supports tasks such as semantic/instance segmentation, referring segmentation, captioning, retrieval, and object classification on raw point clouds via shared query modeling and dynamic task routing. Similarly, UniSeg3D~\cite{UniSeg3D} presents a unified architecture for six segmentation tasks, including open-vocabulary and referring segmentation, using shared encoders and decoders, and enhancing performance through inter-task knowledge distillation and contrastive learning.

\subsubsection{Weakly Supervised/Unsupervised Methods}

To reduce reliance on manual labeling, recent studies explore limited-supervision settings. 3D-REST~\cite{3DResT} addresses the semi-supervised setting by generating high-quality pseudo-labels and reweighting low-confidence predictions based on reliability, enabling effective training with fewer annotations. MEN~\cite{MEN} tackles the weakly supervised setting without mask labels by integrating multimodal cues, \eg, global context, fine-grained attributes, and category priors, to guide accurate segmentation.

\vspace{-1mm}
\subsection{Referring 3D Gaussian Splatting Segmentation}
\label{sec7.2}
To advance research in Referring 3D Gaussian Splatting Segmentation (R3DGS), ReferSplat~\cite{Ref-LERF} introduces a framework that links 3D Gaussians with language via spatially aware referring fields, and proposes Ref-LERF, the first dedicated dataset for this task. Each Gaussian is assigned a learnable feature vector that interacts with textual queries to produce segmentation masks through similarity-based rendering.

Despite recent progress, several challenges remain in R3DGS. First, existing method~\cite{Ref-LERF} is limited to static scenes and lack temporal modeling. Integrating 4D representations, \eg, 4D Gaussian Splatting~\cite{wu20244d}, could enable reasoning in dynamic environments. Second, current method emphasizes segmentation but overlook fine-grained grounding, such as precise localization and size estimation, which are critical for spatial reasoning. Finally, the limited scale and diversity of existing dataset restrict generalization. Advancing the field requires the development of large-scale, diverse datasets to bridge the gap with more mature 2D counterparts.

%% file: Secs/7_GREx.tex
\section{Generalized Referring Expression}
\label{sec:grex}

Generalized Referring Expression tasks (GREx)~\cite{GREx,gres,grec,3d-gres,Ground-V} aim to segment or detect an arbitrary number of target objects in visual scenes based on free-form referring expressions. This flexibility makes GREx more applicable to real-world scenarios and highlights it as a promising direction for future research.

\noindent$\bullet$~{\textbf{Generalized Referring Expression Segmentation (GRES).}} Liu and Ding~\etal~\cite{gres} first introduce the GRES task, which extends classic RES to further support multi-target and no-target expressions. They also propose ReLA, a region-based framework that segments images into semantically meaningful sub-instance regions and explicitly models region–region and region–language dependencies. Following this formulation, numerous works~\cite{gres,dmmi,gsva,r-ris,group-res,r2vos,lqmformer,li2024bring,nguyen2024instance,cohd} have advanced GRES toward more practical and generalizable scenarios. Shah~\etal~\cite{lqmformer} identify a limitation in some GRES methods, including ReLA, termed query collapse, where all queries produce identical mask predictions due to traditional RES’s single-mask constraint. To mitigate this, they propose dynamic query generation conditioned on expressions, along with a regularization strategy to diversify query representations. DMMI~\cite{dmmi} introduces a dual-branch decoder enabling bidirectional interaction: one branch guides visual localization using text, while the other reconstructs expressions from visual features to enforce semantic consistency. GSVA~\cite{gsva} extends the reasoning segmentation paradigm of LISA~\cite{lisa} to the GRES setting by leveraging MLLMs and adapting [\texttt{SEG}] tokens for multi-target references. It further introduces a [\texttt{REJ}] token to to explicitly handle no-target cases.

\noindent$\bullet$~{\textbf{Generalized Referring Expression Comprehension (GREC).}} He and Ding \etal~\cite{grec} introduce the GREC task, an extension of classic REC that allows expressions to refer to any number of target objects. They also propose new evaluation metrics tailored for this setting. As a representative method, HieA2G~\cite{hiea2g} presents a hierarchical multimodal semantic alignment framework that facilitates cross-modal interactions at word-object, phrase-object, and text-image levels. To accommodate varying numbers of targets, it incorporates an adaptive counting mechanism.

\noindent$\bullet$~{\textbf{\!3D Generalized Referring Expression Segmentation (3D-GRES).}}
3D-GRES~\cite{3d-gres} extends GRES~\cite{gres} to the 3D domain, aiming to segment an arbitrary number of target objects in point cloud scenes. The framework employs text-guided query generation and optimization to enable effective interaction between queries, point cloud features, and language, while maintaining semantic consistency across multiple targets.

\noindent$\bullet$~{\textbf{\!3D\! Generalized Referring Expression Comprehension (3D-GREC).}} Zhang~\etal~\cite{multi3drefer} introduce the first 3D-GREC dataset, Multi3DRefer. They propose a CLIP-based method with multi-modal contrastive learning to enable online rendering of proposal objects for generating 2D visual cues. 3D DOG~\cite{3d_dog} handles paragraph-level grounding of multiple objects, while GNL3D~\cite{gnl3d} proposes group-wise grounding across related 3D scenes with flexible target counts.

%% file: Secs/8_Applications.tex
\vspace{-1mm}
\section{Related Tasks and Applications}
\label{sec:related}

\noindent$\bullet$~\!{\textbf{Referring Expression Comprehension (REC).}}
REC~\cite{RefCOCO&RefCOCO+&RefCOCO(g),mattnet,nagaraja2016modeling,mcn,lgran} aims to detect a target object in an image based on a text referring expression by predicting a bounding box around the described object. Unlike RES, REC focuses on target object detection by bounding box rather than pixel-level segmentation.

\noindent$\bullet$~\!{\textbf{3D Referring Expression Comprehension (3D-REC).}}
3D-REC~\cite{scanrefer,referit3d,ffl-3dog,eda} aims to detect a target object in a 3D scene based on a text referring expression. This task takes a 3D point cloud and a referring expression as input, and outputting a 3D bounding box around the described object.

\noindent$\bullet$~\!{\textbf{Referring Video Object Tracking (RVOT).}}
RVOT~\cite{vlttt,decoupletnl,lasot} aims to track a single target object throughout a video based on a text referring expression. The model takes as input a video and a natural language description, and produces a trajectory of bounding boxes for the target object across frames.

\noindent$\bullet${~\!\textbf{Referring Multi-Object Tracking (RMOT).}} 
RMOT~\cite{transrmot,ikun} tracks multiple objects based on text expressions. TransRMOT~\cite{transrmot} uses decoupled object queries for detection and tracking, while iKUN~\cite{ikun} adopts a two-stage framework with a language tracker and a Neural Kalman Filter for adaptive tracking.

\noindent$\bullet$~\!{\textbf{Referring Expression Generation (REG).}} REG~\cite{liu2020attribute,nagaraja2016modeling} aims to generate unambiguous textual referring expressions that uniquely identify specific objects in a visual scene. Some works~\cite{m3att,lang2seg} leverage the synergies between RES and REG through multi-task learning approaches.

\noindent$\bullet$~\!{\textbf{Phrase Grounding (PG).}} PG~\cite{Flickr30k,gupta2020contrastive} aims to ground each entity mentioned by a noun phrase in an image caption to its corresponding region in the image.

\noindent$\bullet$~\!{\textbf{Panoptic Narrative Grounding (PNG).}} PNG~\cite{gonzalez2021panoptic,gonzalez2023piglet} focuses on segmenting both foreground objects and background stuff that are described in detailed narrative captions of an image. While similar to RES, PNG requires the model to identify and segment entities mentioned throughout the entire caption text.

\begin{figure}[t]
	\centering
	\includegraphics[width=1\linewidth]{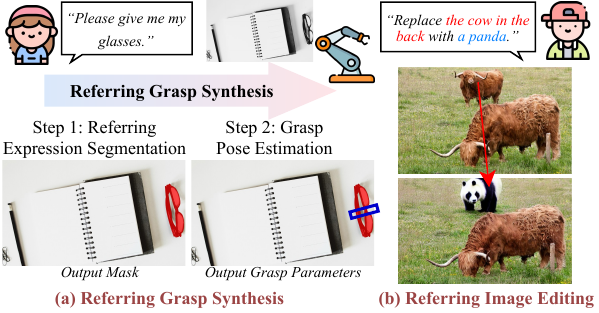}
        \vspace{-6mm}
	\caption{{Illustration of Several Applications.} }\label{fig:application}
    \vspace{-1\baselineskip}
\end{figure}

\noindent$\bullet${~\!\textbf{Referring Expression Counting.}}
This task \cite{dai2024referring,countgd} aims to count object subsets specified by free-form expressions, requiring fine-grained grounding (\eg, “\textit{green grapes}” \vs “\textit{purple grapes}”).

\noindent$\bullet${~\!\textbf{Referring Grasp Synthesis (RGS).}} RGS \cite{ingress,crog} predicts grasp poses for objects described by language, combining vision-language grounding with robotic control. It typically follows a two-stage pipeline: grounding and grasp prediction, see \cref{fig:application}(a).

\noindent$\bullet$~\!{\textbf{Referring Image Editing (RIE).}} RIE~\cite{rie} aims to identify and modify specific objects in an image based on referring expression, see \cref{fig:application}(b). Unlike general image editing~\cite{image_editing_survey}, which often applies global changes, or region-based editing that relies on masks or bounding boxes, RIE focuses on editing objects grounded by referring expressions. This ensures precise region-level editing and also provides flexible and natural user interaction.

%% file: Secs/10_Conclusion.tex
\section{Conclusion and Discussion}
\label{sec:conclusion}

This survey presents a comprehensive overview of multimodal referring segmentation across image, video, and 3D domains. We unify the taxonomy of task settings, summarize a meta-architecture, and analyze representative methods and benchmarks from classic RES and RVOS to emerging tasks like GRES, OmniAVS, and reasoning segmentation, offering a holistic view of the field’s evolution. Recent trends reflect a clear shift toward more general, human-centric, and reasoning-driven models:\\
$\bullet$~\textbf{Omnimodal Understanding.} OmniAVS demonstrates the need for models to flexibly interpret text, speech, sound, and visual cues. This reflects a broader trend toward promptable and generalist models capable of handling diverse multimodal inputs.
$\bullet$~\textbf{Generalization for Practical Scenarios.}
Generalized Referring Expression Segmentation (GRES) extends classical settings by supporting multi-target and no-target expressions. It promotes open-world segmentation and robustness to ambiguous language.\\
$\bullet$~\textbf{Motion-Centric Video Understanding.} MeViS pushes the boundary of RVOS by introducing motion-centric expressions and complex temporal dependencies. It encourages models to perform holistic video reasoning and dynamic object tracking.\\
$\bullet$~\textbf{Foundation Models and Reasoning Segmentation.}
Large vision and language models, such as SAM/SAM2 and MLLMs like LLaVA, have reshaped the landscape of referring segmentation. They enable prompt-based interfaces and spark progress in reasoning segmentation, where models must interpret abstract, indirect, or knowledge-intensive instructions.

Looking ahead, several research directions remain open:
1) Developing generalist segmentation agents that scale across modalities and tasks;
2) Enabling deeper commonsense and temporal reasoning under limited supervision;
3) Enhancing model robustness, efficiency, and interpretability for real-world deployment; and
4) Establishing unified benchmarks to evaluate cross-modal, open-world, and reasoning capabilities.

%% file: Secs/9_Performance_Comparison.tex
\renewcommand{\thesection}{A}
\renewcommand{\thesubsection}{A.\arabic{subsection}}

\section*{Appendix: Performance Comparison}
\label{sec:performance_comparison}

In this section, we present performance comparison of multimodal referring segmentation methods. For each field, the most widely used datasets, as outlined in \cref{sec:background}, are selected for benchmarking. Given the large volume of publications in this area, we selectively report results from representative methods published in top-tier conferences and journals.

\subsection{RES Performance Benchmarking}
\label{RESBench}
\input{Tabs/image.tex}

\input{Tabs/rvos.tex}
\textbf{Evaluation Metrics.}
For Referring Expression Segmentation (RES) evaluation, four primary metrics are commonly used:

\noindent$\bullet$~\textbf{Intersection over Union (IoU)} measures the overlap between predicted ($M_p$) and ground truth ($M_{gt}$) masks: $\text{IoU} = \text{area}(M_p \cap M_{gt})/\text{area}(M_p \cup M_{gt})$. It serves as a fundamental metric for evaluating model accuracy in identifying and segmenting regions.

\noindent$\bullet$~\textbf{Mean Intersection over Union (mIoU)} is defined by the average of all per-image Intersection-over-Unions (IoUs): $\text{mIoU} = \sum_{i=1}^{N}\text{IoU}_i / N$, where $N$ is the total number of data samples. This metric provides a comprehensive evaluation of overall segmentation performance across the entire dataset.

\noindent$\bullet$~\textbf{Cumulative Intersection over Union (cIoU)} is defined by the cumulative intersection over the cumulative union: $\text{cIoU} = {\sum_{i=1}^{N}\text{area}(M_p^i \cap M_{gt}^i)} / {\sum_{i=1}^{N}\text{area}(M_p^i \cup M_{gt}^i)}$. It is worth noting that cIoU is highly biased toward large-area objects and tends to fluctuate significantly, which can affect the reliability of performance assessments.

\noindent$\bullet$~\textbf{Precision@X (Pr@X)} measures the percentage of predictions that achieve an IoU score higher than a given IoU threshold X.

\textbf{RES Benchmark Results.}
As shown in \cref{tab:res}, we present the performance comparison of recent {RES} methods on three widely-used benchmarks: RefCOCO~\cite{ReferItGame}, RefCOCO+~\cite{ReferItGame}, and RefCOCOg~\cite{mao2016generation}.
The field has witnessed substantial progress since 2016, with recent methods achieving remarkable improvements across all datasets.
For conventional fully-supervised methods,
OneRef-L~\cite{oneref} achieves the best performance among conventional fully-supervised methods with 75.68\% mIoU on RefCOCOg validation set and 76.82\% mIoU on the test set, followed closely by UNINEXT~\cite{UNINEXT} with 74.70\% and 76.40\% mIoU respectively. These results demonstrate remarkable progress compared to early methods like LSTM-CNN~\cite{lstm-cnn} from 2016, which achieved only 34.06\% mIoU on RefCOCOg validation set.
For Weakly/Semi-Supervised, Zero-shot and Generalist Methods, ADDP~\cite{addp} achieves the best performance on RefCOCOg with 59.05\% mIoU on the validation set and 59.60\% on the test set, outperforming several conventional fully-supervised methods despite using less supervision.
Among MLLM-based methods, SAM4MLLM~\cite{sam4mllm} achieves the best performance with 75.50\% mIoU on RefCOCOg val set and 76.40\% on the test set, followed by POPEN~\cite{POPEN} with 75.40\% and 75.60\%, respectively, demonstrating the effectiveness of integrating MLLMs into referring segmentation.

\subsection{RVOS Performance Benchmarking}
\label{RVOSBench}

\textbf{Evaluation Metrics.}
Three metrics are frequently used to in Referring Video-Object Segmentation (RVOS) evaluation:

\noindent$\bullet$~\textbf{Region Jaccard $\mathcal{J}$} is calculated by the intersection-over-union (IoU) between the predicted segmentation mask $M_{p}$ and the ground-truth $M_{gt}$:
$\mathcal{J} = {|M_{p} \cap  M_{gt}}|/|{M_{p}\cup M_{gt}|}$,
which computes the number of pixels of the intersection between $M_{p}$ and $M_{gt}$, and divides it by the size of the union.

\noindent$\bullet$~\textbf{Boundary Accuracy $\mathcal{F}$} is the harmonic mean of the boundary precision $\text{P}_c$ and recall $\text{R}_c$. The value of $\mathcal{F}$ reflects how well the predicted contours match the ground-truth contours. Usually, the value of $\text{P}_c$ and $\text{R}_c$ can be computed via bipartite graph matching~\cite{martin2004learning}, then the boundary accuracy $\mathcal{F}$ can be computed as: $\mathcal{F} = {2 \text{P}_c \text{R}_c}/({\text{P}_c + \text{R}_c})$.

\noindent$\bullet$~\textbf{$\mathcal{J}\&\mathcal{F}$} is computed as the average of region similarity and contour accuracy: $\mathcal{J} \& \mathcal{F} = (\mathcal{J} + \mathcal{F})/2$, providing a comprehensive evaluation of both region and boundary accuracy.

\textbf{RVOS Benchmark Results.}
As shown in~\cref{tab:rvos}, we present the performance comparison of recent {RVOS} methods on three widely-used benchmarks: MeViS~\cite{MeViS}, Ref-YouTube-VOS~\cite{Refer-Youtube-VOS}, and Ref-DAVIS$_{17}$~\cite{Refer-DAVIS}. The field has witnessed substantial progress since 2020, with recent methods like GLUS~\cite{GLUS} and VRS-HQ~\cite{vrs-hq} achieving remarkable improvements on all benchmarks. On the MeViS dataset, which focuses on temporal motion understanding, GLUS achieves the best performance with 51.30\% $\mathcal{J}$\&$\mathcal{F}$, followed closely by VRS-HQ with 50.90\%. For Ref-YouTube-VOS, VRS-HQ leads with 71.00\% $\mathcal{J}$\&$\mathcal{F}$, significantly outperforming earlier methods like URVOS~\cite{Refer-Youtube-VOS} (47.23\%). On Ref-DAVIS$_{17}$, VRS-HQ also demonstrates superior performance with 74.40\% $\mathcal{J}$\&$\mathcal{F}$, showing substantial improvement over previous state-of-the-art methods.

\subsection{R-AVS Performance Benchmarking}
\label{RAVSBench}
\input{Tabs/avs.tex}
\textbf{Evaluation Metrics.}
Region Jaccard $\mathcal{J}$, boundary accuracy $\mathcal{F}$ are also widely adopted for Audio-Visual Segmentation (AVS) performance evaluation.

\textbf{AVS Benchmark Results.}
As shown in~\cref{tab:avs}, we present the performance comparison of recent {AVS} methods on AVSBench~\cite{avsbench} (S4 and MS3 subsets) and AVSBench-Semantic~\cite{avsbench_semantic} (AVSS) using $\mathcal{J}$ and $\mathcal{F}$ metrics. The field has seen significant progress since 2022, with recent methods like DDESeg~\cite{DDESeg} and RAVS~\cite{RAVS} achieving remarkable improvements on all benchmarks. DDESeg achieves the best performance on S4 with 92.40\% $\mathcal{J}$ and 95.90\% $\mathcal{F}$, as well as on MS3 with 72.30\% $\mathcal{J}$ and 83.40\% $\mathcal{F}$. For the more challenging AVSS benchmark, DDESeg also leads with 63.40\% $\mathcal{J}$ and 72.30\% $\mathcal{F}$, demonstrating the effectiveness in audio-visual semantic segmentation task.

\textbf{Ref-AVS Benchmark Results.}
As shown in~\cref{tab:refavs}, we present the performance comparison of recent {Ref-AVS} methods on the Ref-AVS~\cite{refavs} dataset using $\mathcal{J}$ and $\mathcal{F}$ metrics across different scenarios. The field has seen remarkable progress since 2022, with recent methods like OmniAVS~\cite{OmniAVS} achieving the best overall performance with 51.7\% $\mathcal{J}$ and 58.7\% $\mathcal{F}$ on the Seen set, and 58.3\% $\mathcal{J}$ and 65.1\% $\mathcal{F}$ on the Unseen set. TSAM~\cite{TSAM} and SAM2-LOVE~\cite{SAM2-LOVE} also demonstrate competitive results, with TSAM achieving 43.4\% $\mathcal{J}$ and 56.8\% $\mathcal{F}$ on Seen, and SAM2-LOVE reaching 66.5\% $\mathcal{J}$ and 72.3\% $\mathcal{F}$ on Unseen scenarios. These results highlight the significant advancement in audio-visual referring segmentation capabilities, substantially outperforming earlier methods like AVSBench~\cite{avsbench} and ReferFormer~\cite{referformer}.

\textbf{OmniAVS Benchmark Results.}
As shown in~\cref{tab:omniavs}, we present the performance comparison of recent methods on the {OmniAVS}~\cite{OmniAVS} dataset using $\mathcal{J}$\&$\mathcal{F}$ metric across eight different splits and METEOR for caption evaluation. OISA-1B~\cite{OmniAVS} achieves the best overall performance with 41.1\% $\mathcal{J}$\&$\mathcal{F}$ across all splits and 21.7 METEOR score, demonstrating significant improvements over previous methods. MUTR~\cite{mutr} and LISA-13B~\cite{lisa} also show competitive results with 32.3\% and 36.1\% overall performance respectively. The results across different splits show varying levels of difficulty, with splits VII and VIII generally achieving higher performance (\eg, OISA-1B's 52.6\% on VII) compared to splits III and IV (\eg, OISA-1B's 34.9\% on III), indicating the heterogeneous nature of the benchmark and the challenges in multi-modal audio-visual understanding.

\subsection{3D-RES Performance Benchmarking}
\label{3DRESBench}
\textbf{Evaluation Metrics.}
mIoU and Acc@X (\ie, Pr@X) are adopted for evaluating 3D-RES performance.
\input{Tabs/3d.tex}

\textbf{Benchmark Results.}
As shown in~\cref{tab:3dres}, we present the performance comparison of recent \textbf{3D-RES} methods on ScanRefer~\cite{scanrefer}. The evaluation is conducted on both unique and multiple reference scenarios, with unique references constituting approximately 19\% of the dataset and multiple references making up the remaining 81\%. IPDN~\cite{ipdn} achieves the best overall performance with 60.60\% Acc@0.25 and 54.90\% Acc@0.5, demonstrating significant improvements over earlier methods. RG-SAN~\cite{Rg-san}, despite being weakly-supervised, shows competitive results with 61.70\% Acc@0.25, though its Acc@0.5 performance (44.90\%) is lower than fully-supervised methods like IPDN~\cite{ipdn} and MDIN~\cite{3d-gres}. Most methods perform substantially better on unique references than multiple references, highlighting the challenge of disambiguating between similar objects in 3D scenes.

\subsection{GREx Performance Benchmarking}
\label{GRExBench}
\input{Tabs/grex.tex}
\textbf{Evaluation Metrics.}
In addition to cIoU, the evaluation metrics for GRES and GREC also include:

\noindent$\bullet$~\textbf{Generalized IoU (gIoU): } The widely-used cIoU tends to favor larger objects. Since multi-target samples have larger foreground areas in GRES, this bias can significantly impact the evaluation results. Similar to mean IoU, gIoU calculates the mean value of per-image IoU over all samples. For no-target samples, the IoU values of true positive no-target samples are regarded as 1, while IoU values of false negative samples are treated as 0.

\noindent$\bullet$~\textbf{N-acc. and T-acc.:} These two metrics to assess model performance on no-target identification. N-acc. (No-target accuracy) evaluates how well a model identifies samples without targets by computing the ratio of true positives, \ie, predictions with no foreground pixels for no-target samples, to total no-target samples: N-acc. = $\text{TP}/({\text{TP}+\text{FN})}$. Meanwhile, T-acc. (Target accuracy) measures the model's ability to avoid misclassifying target-containing samples as no-target samples: T-acc. = ${\text{TN}}/({\text{TN}+\text{FP})}$.

\noindent$\bullet$~\textbf{Precision@($\mathrm{F_1}$=1, IoU$>$0.5):} This metric computes the percentage of samples that achieve an $\mathrm{F_1}$ score of 1 with the IoU threshold set to 0.5. Given a sample and its predicted/ground-truth bounding boxes, a predicted bounding box is regarded as a TP if it has a matched (IoU$>$0.5) ground-truth bounding box. When multiple predicted bounding boxes match one ground-truth bounding box, only the one with the highest IoU is considered TP while others are FP. Ground-truth bounding boxes having no matched predicted bounding box are FN while predicted bounding boxes having no matched ground-truth are FP. The $\mathrm{F_1}$ score is calculated by $\mathrm{F_1} = \frac{2TP}{2TP+FN+FP}$. A sample is considered successfully predicted if the $\mathrm{F_1}$ score equals 1. For no-target samples, the $\mathrm{F_1}$ score is regarded as 1 if there is no predicted bounding box, otherwise 0.

\textbf{GRES Benchmark Results.}
As shown in~\cref{tab:gres}, we present the performance comparison of recent \textbf{GRES} methods on the gRefCOCO~\cite{gres} dataset using cIoU and mIoU metrics. The field has seen significant progress since the GRES~\cite{gres} task was proposed, with recent methods like InstAlign~\cite{nguyen2024instance} achieving the best performance with 68.94\% cIoU and 74.34\% mIoU on the validation set. GSVA~\cite{gsva} also demonstrates strong performance, reaching 66.38\% cIoU and 70.04\% mIoU on the validation set. On the testA and testB sets, InstAlign continues to lead with 73.22\% cIoU, 74.51\% mIoU and 63.88\% cIoU, 65.74\% mIoU respectively. These improvements highlight the rapid advancement in generalized referring expression segmentation capabilities and challenges, significantly outperforming earlier methods like MattNet~\cite{mattnet} and VLT~\cite{vlt} designed for only single-object tasks.

\textbf{GREC Benchmark Results.}
As shown in~\cref{GREC}, we present the performance comparison of recent \textbf{GREC} methods on the gRefCOCO~\cite{gres} dataset using Pr@($\mathrm{F_1}$=1, IoU$\geq$0.5) and N-acc. metrics. The field has witnessed significant advancement since the introduction of GREC task, with recent methods like HieA2G~\cite{hiea2g} achieving the best overall performance with 67.8\% Pr and 60.3\% N-acc. on the validation set. SimVG~\cite{SimVG} also demonstrates competitive results, reaching 62.1\% Pr and 54.7\% N-acc. on validation. On the testA set, SimVG leads with 64.6\% Pr, while NGDINO~\cite{RefDrone} achieves the highest N-acc. at 83.2\%. For testB evaluation, HieA2G maintains strong performance with 56.5\% Pr and 56.0\% N-acc.

\subsection{ReasonSeg Performance Benchmarking}
\label{ReasonBench}
\input{Tabs/reasonseg.tex}
\textbf{Evaluation Metrics.}
mIoU and cIoU are adopted for evaluating Reasoning Segmentation performance.

\textbf{ReasonSeg Benchmark Results.}
As shown in~\cref{tab:reasonseg}, we present the performance comparison of recent Reasoning Segmentation (ReasonSeg) methods on the ReasonSeg~\cite{lisa} dataset using gIoU and cIoU metrics. The results demonstrate a clear performance gap between methods with and without Multimodal Large Language Models (MLLMs). Traditional methods without MLLMs, such as ReLA~\cite{gres}, X-Decoder~\cite{x-decoder}, and OVSeg~\cite{OvSeg}, achieve modest performance with gIoU scores ranging from 22.4\% to 28.5\% on the validation set. In contrast, MLLM-based approaches show substantial improvements, with CoReS~\cite{CoReS} achieving the best performance at 68.1\% gIoU on the validation set, followed by VisionReasoner~\cite{VisionReasoner} at 66.3\%. On the test set, CoReS leads with 65.5\% gIoU, followed by VisionReasoner at 63.6\%. When comparing 7B parameter MLLM models for fair comparison, VisionReasoner achieves the best performance with 66.3\% gIoU on validation and 63.6\% on test, followed by SAM-R1~\cite{SAM-R1} and PixelThink~\cite{PixelThink} (both 60.2\% gIoU on test). These results highlight the significant advantage of leveraging MLLMs for complex reasoning tasks in segmentation, with fine-tuned models consistently outperforming their non-MLLM counterparts by large margins across both validation and test sets.

%% file: Tabs/image.tex
\begin{table*}[!htp]
  \setlength{\tabcolsep}{13pt}
      \footnotesize
          \caption{\textbf{Performance Comparison on RES benchmarks.} The results are evaluated on RefCOCO~\cite{ReferItGame}, RefCOCO+~\cite{ReferItGame}, and RefCOCOg~\cite{mao2016generation} (UMD partition) using mIoU. $\dagger$: results post-processed with DenseCRF~\cite{DenseCRF}. $*$: methods utilizing 30\% mask annotations and 70\% bounding box annotations. `ft' denotes fine-tuning on referring expression segmentation datasets. $\ddag$ indicates methods using only 5\% labeled data. UNINEXT~\cite{UNINEXT} utilizes val/test set masks during training, leading to mask information leakage.}
      \vspace{-8pt}
      \centering
      \resizebox{\textwidth}{!}{\begin{tabular}{rcccccccccc}
      \toprule
      \multirow{2}[2]{*}{\textbf{Method}} & \multirow{2}[2]{*}{\textbf{Venue}} & \multicolumn{3}{c}{\textbf{RefCOCO}} & \multicolumn{3}{c}{\textbf{RefCOCO+}} & \multicolumn{2}{c}{\textbf{RefCOCOg}} \\
      \cmidrule(lr){3-5} \cmidrule(lr){6-8} \cmidrule(lr){9-10}
                                  &   & \textbf{val}  & \textbf{test A} & \textbf{test B} & \textbf{val} & \textbf{test A} & \textbf{test B} & \textbf{val} & \textbf{test} \\
      \midrule
    \multicolumn{10}{l}{\textbf{\textit{a. Conventional Fully-supervised methods.}}}        \\
    \rowcolor{cyan!07}
      LSTM-CNN~\cite{lstm-cnn} & \pub{ECCV'16} & - & - & - & - & - & - & 34.06 & - \\
      RMI$^\dagger$~\cite{rmi} & \pub{ICCV'17} & 45.18 & 45.69 & 45.57 & 29.86 & 30.48 & 29.50 & 34.52 & - \\
    \rowcolor{cyan!07}
      DMN~\cite{dmn} &\pub{ECCV'18}  & 49.78 & 54.83 & 45.13 & 38.88 & 44.22 & 32.29 & - & - \\
      KWA~\cite{kwa} & \pub{ECCV'18} & - & - & - & - & - & - & 36.92 & - \\
    \rowcolor{cyan!07}
      RRN$^\dagger$~\cite{rrn} & \pub{CVPR'18} & 55.33 & 57.26 & 53.93 & 39.75 & 42.15 & 36.11 & 36.45 & - \\
      MAttNet~\cite{mattnet} & \pub{CVPR'18} & 56.51 & 62.37 & 51.70 & 46.67 & 52.39 & 40.08 & 47.64 & 48.61 \\
    \rowcolor{cyan!07}
      CMSA$^\dagger$~\cite{cmsa} & \pub{CVPR'19} & 58.32 & 60.61 & 55.09 & 43.76 & 47.60 & 37.89 & 39.98 & - \\
      LSCM$^\dagger$~\cite{lscm} & \pub{ECCV'20} & 61.47 & 64.99 & 59.55 & 49.34 & 53.12 & 43.50 & - & - \\
    \rowcolor{cyan!07}
      ConvLSTM$^\dagger$~\cite{convlstm} &\pub{TMM'20} & 59.04 & 60.74 & 56.73 & 44.54 & 47.92 & 39.73 & 41.77 & - \\
      MCN~\cite{mcn} & \pub{CVPR'20} & 62.44 & 64.20 & 59.71 & 50.62 & 54.99 & 44.69 & 49.22 & 49.40 \\
    \rowcolor{cyan!07}
      BRINet~\cite{brinet} & \pub{CVPR'20} & 60.98 & 62.99 & 59.21 & 48.17 & 52.32 & 42.11 & - & - \\
      STEP~(5-fold)~\cite{step} & \pub{NeurIPS'21} & 60.04 & 63.46 & 57.97 & 48.19 & 52.33 & 40.41 & 46.40 & - \\
    \rowcolor{cyan!07}
      Referring Transformer~\cite{referring_transformer} & \pub{NeurIPS'21} & 74.34 & 76.77 & 70.87 & 66.75 & 70.58 & 59.40 & 66.63 & 67.39 \\
      EFN~\cite{efn} & \pub{CVPR'21} & 62.76 & 65.69 & 59.67 & 51.50 & 55.24 & 43.01 & 51.93 & - \\
    \rowcolor{cyan!07}
      BUSNet~\cite{busnet} & \pub{CVPR'21} & 63.27 & 66.41 & 61.39 & 51.76 & 56.87 & 44.13 & - & - \\
      LTS~\cite{lts} & \pub{CVPR'21} & 65.43 & 67.76 & 63.08 & 54.21 & 58.32 & 48.02 & 54.40 & 54.25 \\
    \rowcolor{cyan!07}
      CMPC$^\dagger$~\cite{cmsa} & \pub{TPAMI'21} & 61.36 & 64.53 & 59.64 & 49.56 & 53.44 & 43.23 & - & - \\

      SeqTR~\cite{seqtr} & \pub{ECCV'22} & 67.26 & 69.79 & 64.12 & 54.14 & 58.93 & 48.19 & 55.67 & 55.64 \\
    \rowcolor{cyan!07}
      CoupAlign~\cite{coupalign} & \pub{NeurIPS'22} & 74.70 & 77.76 & 70.58 & 62.92 & 68.34 & 56.69 & 62.84 & 62.22 \\
      ReSTR~\cite{restr} & \pub{CVPR'22} & 67.22 & 69.30 & 64.45 & 55.78 & 60.44 & 48.27 & - & - \\
    \rowcolor{cyan!07}
      CRIS~\cite{cris} & \pub{CVPR'22} & 70.47 & 73.18 & 66.10 & 62.27 & 68.08 & 53.68 & 59.87 & 60.36 \\
      LAVT~\cite{lavt} & \pub{CVPR'22} & 72.73 & 75.82 & 68.79 & 62.14 & 68.38 & 55.10 & 61.24 & 62.09 \\
    \rowcolor{cyan!07}
      BVPR~\cite{kesen2022modulating} & \pub{CVPR'22} & 67.01 & 69.63 & 63.45 & 55.34 & 60.72 & 47.11 & 55.09 & 55.31 \\
      BKINet~\cite{bkinet} & \pub{TMM'23} & 73.22 & 76.43 & 69.42 & 64.91 & 69.88 & 53.39 & 64.21 & 63.77 \\
    \rowcolor{cyan!07}
      M$^3$Att~\cite{m3att} &\pub{TIP'23}  & 73.60 & 76.23 & 70.36 & 65.34 & 70.50 & 56.98 & 64.92 & 67.37 \\
      ETRIS~\cite{etris} & \pub{ICCV'23} & 70.51 & 73.51 & 66.63 & 60.10 & 66.89 & 50.17 & 59.82 & 59.91 \\
    \rowcolor{cyan!07}
      VPD~\cite{vpd} & \pub{ICCV'23} & 73.25 & - & - & 62.69 & - & - & 61.96 & - \\
      TRIS~\cite{tris} & \pub{ICCV'23} &  41.10 & 48.10 & 31.90 & 31.60 & 31.90 & 30.60 & 39.00 & 39.90 \\
    \rowcolor{cyan!07}
      PolyFormer~\cite{polyformer} & \pub{CVPR'23} & 75.96 & 77.09 & 73.22 & 70.65 & 74.51 & 64.64 & 69.36 & 69.88 \\
      CGFormer~\cite{cgformer} &\pub{CVPR'23}  & 76.93 & 78.70 & 73.32 & 68.56 & 73.76 & 61.72 & 67.57 & 67.83 \\
    \rowcolor{cyan!07}
      VG-LAW~\cite{vg-law} & \pub{CVPR'23} & 75.62 & 77.51 & 72.89 & 66.63 & 70.38 & 58.89 & 65.63 & 66.08 \\
      UNINEXT~\cite{UNINEXT} &\pub{CVPR'23} & \textbf{82.20} & \textbf{83.40} & \textbf{81.30} & 72.50 & 76.40 & 66.20 & 74.70 & \textbf{76.40} \\
    \rowcolor{cyan!07}
      ReLA~\cite{gres} & \pub{CVPR'23} & 73.82 & 76.48 & 70.18 & 66.04 & 71.02 & 57.65 & 65.00 & 65.97 \\
      VLT~\cite{vlt,vlticcv} & \pub{TPAMI'23} & 65.65 & 68.29 & 62.73 & 55.50 & 59.20 & 49.36 & 52.99 & 56.65 \\
    \rowcolor{cyan!07}
      CM-MaskSD~\cite{cm-masksd} & \pub{TMM'24} & 74.89 & 77.54 & 71.28 & 67.47 & 71.80 & 59.91 & 66.53 & 66.63 \\
      ReMamber~\cite{remamber} & \pub{ECCV'24} & 71.60 & 73.30 & 68.40 & 61.60 & 65.80 & 54.00 & 61.10 & 61.20 \\
    \rowcolor{cyan!07}
      Barleria~\cite{barleria} & \pub{ICLR'24} & 72.40 & 75.90 & 68.30 & 65.00 & 70.80 & 56.90 & 63.40 & 63.80 \\
      UniRES~\cite{mres} & \pub{CVPR'24} & 79.20 & 81.60 & 76.60 & \textbf{73.00} & 78.10 & 65.80 & 71.70 & 73.20 \\
    \rowcolor{cyan!07}
      MagNet~\cite{magnet} & \pub{CVPR'24} & 76.55 & 78.27 & 72.15 & 68.10 & 73.64 & 61.81 & 67.79 & 69.29 \\
      LQMFormer~\cite{lqmformer} & \pub{CVPR'24} & 74.16 & 76.82 & 71.04 & 65.91 & 71.84 & 57.59 & 64.73 & 66.04 \\
    \rowcolor{cyan!07}
      Prompt-RIS~\cite{prompt-ris} & \pub{CVPR'24} & 78.10 & 81.21 & 74.64 & 71.13 & 76.60 & 64.25 & 70.47 & 71.29 \\
      OneRef-L~\cite{oneref} & \pub{NeurIPS'24} & 81.26 & 83.06 & 79.45 & 76.60 & \textbf{80.16} & \textbf{72.95} & \textbf{75.68} & 76.82 \\
    \midrule
    \multicolumn{10}{l}{\textbf{\textit{b. Weakly/Semi- Supervised, Zero-shot and Generalist Methods.}}}        \\
    \rowcolor{cyan!07}
    Weakly-RIS~\cite{weakly-ris} & \pub{ICCV'23} & 31.06 & 32.30 & 30.11 & 31.28 & 32.11 & 30.13 & 32.88 & - \\
    SaG~\cite{sag} & \pub{ICCV'23} & 44.60 & 50.10 & 38.40 & 35.50 & 41.10 & 27.60 & - & - \\
    \rowcolor{cyan!07}
    Global-Local CLIP~\cite{global-local-clip} & \pub{CVPR'23}  & 26.20 & 24.94 & 26.56 & 27.80 & 25.64 & 27.84 & 33.52 & 33.67 \\
    X-Decoder~\cite{x-decoder} & \pub{CVPR'23} & - & - & - & - & - & - & 64.60 & - \\
    \rowcolor{cyan!07}
    Partial-RES$^*$~\cite{partial-res} & \pub{CVPR'23} & 66.24 & 68.39 & 63.57 & 54.37 & 58.16 & 47.92 & 54.69 & 54.81 \\
    GTMS~\cite{gtms} & \pub{ECCV'24} & 66.54 & 69.98 & 63.41 & \textbf{57.59} & \textbf{63.46} & 50.32 & 54.52 & 54.75 \\
    \rowcolor{cyan!07}
    SAFARI$^*$~\cite{safari} & \pub{ECCV'24} & 67.04 & 69.17 & 64.23 & 54.98 & 59.31 & 48.26 & 55.72 & 55.83 \\
    SemiRES$^\ddag$~\cite{semires} & \pub{ICML'24} & 61.31 & 66.64 & 55.94 & 47.00 & 54.42 & 38.74 & 47.61 & 50.11 \\
    \rowcolor{cyan!07}
    PPT \cite{ppt} & \pub{CVPR'24} &  46.76  &   45.33 & 46.28 & 45.34 & 45.84 & 44.77 & 42.97 &  - \\
    SEEM~\cite{seem} & \pub{NeurIPS'24} & - & - & - & - & - & - & \textbf{65.70} & - \\
    \rowcolor{cyan!07}
    PCNet~\cite{pcnet} & \pub{NeurIPS'24} & 52.20 & 58.40 & 42.10 & 47.90 & 56.50 & 36.20 & 46.80 & 46.90 \\
    HybridGL~\cite{HybridGL} & \pub{CVPR'25} & 49.48 & 53.37 & 45.19 & 43.40 & 49.13 & 37.17 & 51.25 & 51.59 \\
    \rowcolor{cyan!07}
    ADDP~\cite{addp} & \pub{ICLR'25} & \textbf{69.14} & \textbf{70.27} & \textbf{67.46} & 57.58 & 61.65 & \textbf{51.76} & 59.05 & \textbf{59.60} \\

    \midrule
    \multicolumn{10}{l}{\textbf{\textit{c. MLLMs based Methods.}}}        \\
    \rowcolor{cyan!07}
      LISA-7B~(ft)~\cite{lisa} & \pub{CVPR'24} & 74.90 & 79.10 & 72.30 & 65.10 & 70.80 & 58.10 & 67.90 & 70.60 \\
      PixelLm-7B~\cite{pixellm} & \pub{CVPR'24} & 73.00 & 76.50 & 68.20 & 66.30 & 71.70 & 58.30 & 69.30 & 70.50 \\
    \rowcolor{cyan!07}
      SESAME-7B~\cite{SESAME} & \pub{CVPR'24} & 74.70 & - & - & 64.90 & - & - & 66.10 & - \\
      AnyRef-7B~(ft)~\cite{anyref} & \pub{CVPR'24} & 76.90 & 79.90 & 74.20 & 70.30 & 73.50 & 61.80 & 70.00 & 70.70 \\
    \rowcolor{cyan!07}
      PerceptionGPT-13B~\cite{PerceptionGPT} &\pub{CVPR'24} & 75.30 & 79.10 & 72.10 & 68.90 & 74.00 & 61.90 & 70.70 & 71.90 \\
      GSVA-7B~(ft)~\cite{gsva} & \pub{CVPR'24} & 77.20 & 78.90 & 73.50 & 65.90 & 69.60 & 59.80 & 72.70 & 73.30 \\
    \rowcolor{cyan!07}
      GLaMM-7B~\cite{glamm} & \pub{CVPR'24} & 79.50 & 83.20 & 76.90 & 72.60 & 78.70 & 64.60 & 74.20 & 74.90 \\
      CoReS-7B~\cite{CoReS} & \pub{ECCV'24} & 76.00 & 78.60 & 72.50 & 65.10 & 70.00 & 58.60 & 69.00 & 70.70 \\
    \rowcolor{cyan!07}
      SAM4MLLM-8B~\cite{sam4mllm} & \pub{ECCV'24} & 79.80 & \textbf{82.70} & \textbf{74.70} & \textbf{74.60} & \textbf{80.00} & \textbf{67.20} & \textbf{75.50} & \textbf{76.40} \\
      M$^2$SA-13B~\cite{MMR} & \pub{ICLR'25} & 74.60 & 77.60 & 71.00 & 64.00 & 68.10 & 57.60 & 69.00 & 69.30 \\
    \rowcolor{cyan!07}
      SegLLM-7B~\cite{segllm} &\pub{ICLR'25} & \textbf{80.20} & 81.50 & 75.40 & 70.30 & 73.00 & 62.50 & 72.60 & 73.60 \\
      READ-7B~\cite{READ} & \pub{CVPR'25} & 78.10 & 80.20 & 73.20 & 68.40 & 73.70 & 60.40 & 70.10 & 71.40 \\
    \rowcolor{cyan!07}
      POPEN-7B~(ft)~\cite{POPEN} & \pub{CVPR'25} & 79.30 & 82.00 & 74.10 & 73.10 & 77.00 & 65.10 & 75.40 & 75.60 \\
      \bottomrule
      \end{tabular}%
      }
      \label{tab:res}%
\end{table*}%

%% file: Tabs/rvos.tex
\begin{table*}[!t]
    \centering
    \caption{\textbf{Performance Comparison on RVOS Benchmarks.} The results are evaluated on MeViS~\cite{MeViS}, Ref-YouTube-VOS~\cite{Refer-Youtube-VOS}, and Ref-DAVIS$_{17}$~\cite{Refer-DAVIS} datasets with $\mathcal{J}$ and $\mathcal{F}$ metrics.}
    \vspace{-3mm}
    \setlength{\tabcolsep}{10.5pt}
    \begin{tabular}{rcccccccccc}
    \toprule
    \multirow{2}{*}{\textbf{Method}} & \multirow{2}{*}{\textbf{Venue}} & \multicolumn{3}{c}{\textbf{MeViS}} & \multicolumn{3}{c}{\textbf{Ref-YouTube-VOS}} & \multicolumn{3}{c}{\textbf{Ref-DAVIS$_{17}$}} \\
    \cmidrule(lr){3-5} \cmidrule(lr){6-8} \cmidrule(lr){9-11}
    & & \textbf{$\mathcal{J}$\&$\mathcal{F}$} & \textbf{$\mathcal{J}$} & \textbf{$\mathcal{F}$} & \textbf{$\mathcal{J}$\&$\mathcal{F}$} & \textbf{$\mathcal{J}$} & \textbf{$\mathcal{F}$} & \textbf{$\mathcal{J}$\&$\mathcal{F}$} & \textbf{$\mathcal{J}$} & \textbf{$\mathcal{F}$} \\
    \midrule
    \rowcolor{cyan!07}
    URVOS~\cite{Refer-Youtube-VOS} &\pub{ECCV'20}  & 27.80 & 25.70 & 29.90 & 47.23 & 45.27 & 49.19 & 51.63 & 47.29 & 55.96 \\
    CMPC~\cite{cmpc-pami} & \pub{TPAMI'21} & - & - & - & 47.48 & 45.64 & 49.32 & - & - & - \\
    \rowcolor{cyan!07}
    MLRL~\cite{mlrl} & \pub{CVPR'22} & - & - & - & 49.70 & 48.43 & 50.96 & 57.94 & 53.85 & 62.02 \\
    LBDT~\cite{lbdt} & \pub{CVPR'22} & 29.30 & 27.80 & 30.80 & 49.38 & 48.18 & 50.57 & 54.52 & - & - \\
    \rowcolor{cyan!07}
    MTTR~\cite{mttr} & \pub{CVPR'22} & 30.00 & 28.80 & 31.20 & 55.32 & 54.00 & 56.64 & - & - & - \\
    ReferFormer~\cite{referformer} & \pub{CVPR'22} & 31.00 & 29.80 & 32.20 & 62.90 & 61.30 & 64.60 & 61.10 & 58.10 & 64.10 \\

    \rowcolor{cyan!07}
    EFCMA~\cite{efcma} & \pub{TPAMI'22} & - & - & - & 48.97 & 47.82 & 50.12 & 50.23 & 47.37 & 53.08 \\
    HTML~\cite{html} & \pub{ICCV'23} & - & - & - & 63.40 & 61.50 & 65.20 & 62.10 & 59.20 & 65.10 \\
    \rowcolor{cyan!07}
    SgMg~\cite{sgmg} & \pub{ICCV'23} & - & - & - & 65.70 & 63.90 & 67.40 & 63.30 & 60.60 & 66.00 \\
    TempCD~\cite{tempcd} & \pub{ICCV'23} & - & - & - & 65.80& 63.60 & 68.00 & 64.60 & 61.60 & 67.60 \\
    \rowcolor{cyan!07}
    UniRef++~\cite{uniref++} &\pub{ICCV'23} & - & - & - & 66.90 & 64.80 & 69.00 & 67.20 & 63.40 & 70.90 \\
    OnlineRefer~\cite{onlinerefer} & \pub{ICCV'23} & 32.30 & 31.50 & 33.10 & 63.50 & 61.60 & 65.50 & 64.80 & 61.60 & 67.70 \\
    \rowcolor{cyan!07}
    LMPM~\cite{MeViS} & \pub{ICCV'23} & 37.20 & 34.20 & 40.20 & - & - & - & - & - & - \\

    LASTC~\cite{lastc} & \pub{TPAMI'23} & - & - & - & 49.30 & 48.15 & 50.45 & 54.45 & - & - \\
    \rowcolor{cyan!07}
    Locater~\cite{locater} &\pub{TPAMI'23} & - & - & - & 56.50 & 54.80 & 58.10 & - & - & - \\
    VD-IT~\cite{vd-it} & \pub{ECCV'24} & - & - & - & 66.50 & 64.40 & 68.50 & 69.40 & 66.20 & 72.60 \\
    \rowcolor{cyan!07}
    VISA-13B~\cite{visa} & \pub{ECCV'24} & 44.50 & 41.80 & 47.10 & 63.00 & 61.40 & 64.70 & 70.40 & 67.00 & 73.80 \\

    SOC~\cite{soc} & \pub{NeurIPS'24} & - & - & - & 67.30 & 65.30 & 69.30 & 65.80 & 62.50 & 69.10 \\
    \rowcolor{cyan!07}
    VideoLISA-3.8B~\cite{videolisa} &\pub{NeurIPS'24} & 42.30 & 39.40 & 45.20 & 61.70 & 60.20 & 63.30 & 67.70 & 63.80 & 71.50 \\

    UniVS~\cite{univs} & \pub{CVPR'24} & - & - & - & 58.00 & - & - & 59.40 & - & - \\
    \rowcolor{cyan!07}
    LoSh~\cite{losh} & \pub{CVPR'24} & - & - & - & 67.20 & 65.40 & 69.00 & 64.30 & 61.80 & 66.80 \\
    DsHmp~\cite{dshmp} &\pub{CVPR'24} & 46.40 & 43.00 & 49.80 & 67.10 & 65.00 & 69.10 & 64.90 & 61.70 & 68.10 \\
    \rowcolor{cyan!07}
    ViLLa~\cite{villa} & \pub{ArXiv'24} & 49.40  & 46.50 & 52.30 & 67.50 & 64.60 & 70.40 & 74.30 & 70.60 & 78.00 \\
    SAMWISE~\cite{samwise} & \pub{CVPR'25} & 48.30 & 45.40 & 51.20 & 67.20 & 65.20 & 69.30 & 68.50 & 65.60 & 71.50 \\
    \rowcolor{cyan!07}
    VRS-HQ-13B~\cite{vrs-hq} & \pub{CVPR'25} & 50.90 & 48.00 & 53.70 & 71.00 & 69.00 & 73.10 & 74.40 & \textbf{71.00} & 77.90 \\
    GLUS~\cite{GLUS} & \pub{CVPR'25} & 51.30 & 48.50 & 54.20 & 67.30 & 65.50 & 69.00 & - & - & - \\
    \rowcolor{cyan!07}
    Sa2VA-26B~\cite{sa2va} & \pub{ArXiv'25} & 46.20 & - & - & 70.10 & - & - & \textbf{77.00} & - & - \\
    RGA3-7B~\cite{RGA3} & \pub{ICCV'25} & 50.10 & 47.40 & 52.80 & 68.50 & 66.80 & 70.10 &72.80& 68.30&77.30 \\
    \rowcolor{cyan!07}
    MPG-SAM 2~\cite{mpg-sam_2} & \pub{ICCV'25} &\textbf{53.70}  &\textbf{50.70}  & \textbf{56.70} & \textbf{73.90} & \textbf{71.70} & \textbf{76.10} & 72.40 & 68.80 & \textbf{78.00} \\
    \bottomrule
    \end{tabular}
    \label{tab:rvos}
\end{table*}

%% file: Tabs/avs.tex
\begin{table}[!t]
    \centering
    \caption{\textbf{Performance Comparison on AVS Benchmarks. }
    The results are evaluated on AVSBench~\cite{avsbench} and AVSBench-Semantic~\cite{avsbench_semantic} with $\mathcal{J}$ and $\mathcal{F}$ score metrics. $*$~: weakly-supervised.
    }
    \vspace{-3mm}
    \setlength{\tabcolsep}{3.96pt}
    \resizebox{0.48\textwidth}{!}{
    \begin{tabular}{rccccccc}
    \toprule
    \multirow{2}{*}{\textbf{Method}} & \multirow{2}{*}{\textbf{Venue}} & \multicolumn{2}{c}{\textbf{S4}} & \multicolumn{2}{c}{\textbf{MS3}} & \multicolumn{2}{c}{\textbf{AVSS}} \\
    \cmidrule(lr){3-4} \cmidrule(lr){5-6} \cmidrule(lr){7-8}
    & & \textbf{$\mathcal{J}$} & \textbf{$\mathcal{F}$} & \textbf{$\mathcal{J}$} & \textbf{$\mathcal{F}$} & \textbf{$\mathcal{J}$} & \textbf{$\mathcal{F}$} \\
    \midrule
    \rowcolor{cyan!07}
    AVSBench~\cite{avsbench} & \pub{ECCV'22} & 78.74 & 87.90 & 54.00 & 64.50 & 29.77 & 35.20 \\
    ECMVAE~\cite{ecmvae} & \pub{ICCV'23} & 81.74 & 90.10 & 57.84 & 70.80 & - &- \\
    \rowcolor{cyan!07}
    LAVISH~\cite{lavish} & \pub{CVPR'23} & 80.10 & - & - & - & - & -\\
    PIF~\cite{pif} & \pub{TMM'24} & 81.40 & 90.00 & 58.90 & 70.90 & - & -\\
    \rowcolor{cyan!07}
    BAVS~\cite{bavs} & \pub{TMM'24} & 82.68 & 89.75 & 59.63 & 65.89 & 33.59 & 37.52 \\
    C3N~\cite{c3n} & \pub{TMM'24} & 83.11 & 90.80 & 61.72 & 72.20 & - & -\\
    \rowcolor{cyan!07}
    CPM~\cite{cpm} & \pub{ECCV'24} & 81.37 & 90.47 & 59.80 & 71.00 & 34.53 & 39.57\\
    Stepping Stones~\cite{stepping_stones} & \pub{ECCV'24} & 83.20 & 91.30 & 67.30 & 77.60 & 48.50 & 53.20\\
    \rowcolor{cyan!07}
    TESO~\cite{teso} & \pub{ECCV'24} & 83.27 & 93.30 & 66.02 & 80.10 & 38.96 & 45.10 \\
    WS-AVS$^*$~\cite{ws-avs} & \pub{NeurIPS'24} & 34.13 & 51.76 & 30.85 & 46.87 & - & - \\
    \rowcolor{cyan!07}
    QDFormer~\cite{qdformer} & \pub{CVPR'24} & 79.50 & 88.20 & 61.90 & 66.10 & 53.40 & - \\
    COMBO~\cite{combo} & \pub{CVPR'24} & 84.70 & 91.90 & 59.20 & 71.20 & 42.10 & 46.10\\
    \rowcolor{cyan!07}
    VPO~\cite{vpo} & \pub{CVPR'24} & 85.77 & 92.86 & 62.39 & 73.62 & 44.70 & 57.76\\
    DeepAVFusion~\cite{deepavfusion} & \pub{CVPR'24} & 89.94 & 92.34 & 52.05 & 58.29 & - & -\\
    \rowcolor{cyan!07}
    RAVS~\cite{RAVS} & \pub{CVPR'25} & \textbf{93.10} & 93.80 & 70.60 & 82.10 & 60.80 & 70.60\\
    DDESeg~\cite{DDESeg} & \pub{CVPR'25} & 92.40  & \textbf{95.90}  & \textbf{72.30}  & \textbf{83.40}  & \textbf{63.40}  & \textbf{72.30}  \\
    \bottomrule
    \end{tabular}
    }
    \label{tab:avs}
\end{table}

\begin{table}[t]
    \centering
    \footnotesize
    \setlength\tabcolsep{2.5pt}
    \caption{\textbf{Performance Comparison on Ref-AVS Benchmark.} The results are evaluated on Ref-AVS~\cite{refavs} dataset with $\mathcal{J}$ and $\mathcal{F}$ metrics. $\mathcal{S}$ denotes the square root of the ratio between the predicted mask area and the background area, with lower values indicating better performance.}
    \vspace{-3mm}
    \label{tab:refavs}
    \begin{tabular}{lccccccccc}
        \toprule
        \multirow{2}{*}{\textbf{Method}} & \multirow{2}{*}{\textbf{Venue}} & \multicolumn{2}{c}{\textbf{Seen}} & \multicolumn{2}{c}{\textbf{Unseen}} & \multicolumn{2}{c}{\textbf{Mix(S+U)}} & \multicolumn{2}{c}{\textbf{Null}} \\
        \cmidrule(lr){3-4} \cmidrule(lr){5-6} \cmidrule(lr){7-8} \cmidrule(lr){9-10}
        & & $\mathcal{J}$ & $\mathcal{F}$ & $\mathcal{J}$ & $\mathcal{F}$ & $\mathcal{J}$ & $\mathcal{F}$ & \multicolumn{2}{c}{$\mathcal{S}$ ($\downarrow$)} \\
        \midrule
        \rowcolor{cyan!07}
        AVSBench~\cite{avsbench} & \pub{ECCV'22} & 23.2 & 51.1 & 32.4 & 54.7 & 27.8 & 52.9 & \multicolumn{2}{c}{20.8} \\
        ReferFormer~\cite{referformer} & \pub{CVPR'22} & 31.3 & 50.1 & 30.4 & 48.8 & 30.9 & 49.5 & \multicolumn{2}{c}{17.6} \\
        \rowcolor{cyan!07}
        R2VOS~\cite{r2vos} & \pub{ICCV'23} & 25.0 & 41.0 & 27.9 & 49.8 & 26.5 & 45.4 & \multicolumn{2}{c}{18.3} \\
        AVGSegFormer~\cite{avsegformer} & \pub{AAAI'24} & 33.5 & 47.0 & 36.1 & 50.1 & 34.8 & 48.6 & \multicolumn{2}{c}{17.1} \\
        \rowcolor{cyan!07}
        GAVS~\cite{gavs} & \pub{AAAI'24} & 28.9 & 49.8 & 29.8 & 49.7 & 29.4 & 49.8 & \multicolumn{2}{c}{19.0} \\
        RefAVS~\cite{refavs} & \pub{ECCV'24} & 34.2 & 51.3 & 49.5 & 64.8 & 41.9 & 58.1 & \multicolumn{2}{c}{\ \ \textbf{0.7}} \\
        \rowcolor{cyan!07}
        TSAM~\cite{TSAM} & \pub{CVPR'25} & 43.4 & 56.8 & 54.6 & 66.4 & - & - & \multicolumn{2}{c}{\ \ 1.7} \\
        SAM2-LOVE~\cite{SAM2-LOVE}& \pub{CVPR'25} & 43.5 & 51.9 & \textbf{66.5} & \textbf{72.3} & \textbf{55.0} & \textbf{62.1} & \multicolumn{2}{c}{23.0} \\
        \rowcolor{cyan!07}
        OISA-1B~\cite{OmniAVS} & \pub{ICCV'25} & \textbf{51.7} & \textbf{58.7} & 58.3 & 65.1 & 54.5 & 61.4 & \multicolumn{2}{c}{\ \ 9.8} \\
        \bottomrule
    \end{tabular}
\end{table}

\begin{table}[!t]
    \centering
    \caption{\textbf{Performance Comparison on OmniAVS Benchmark. } The results are evaluated on OmniAVS~\cite{OmniAVS} dataset with $\mathcal{J}\&\mathcal{F}$. \textit{All} is the average result across 8 splits. \textit{MET.}: METEOR.}
    \vspace{-3mm}
    \setlength{\tabcolsep}{3.3pt}
    \resizebox{0.490\textwidth}{!}{
        \begin{tabular}{lccccccccc|c}
            \toprule
            \textbf{Method} & \textbf{All} & \textbf{I} & \textbf{II} & \textbf{III} & \textbf{IV} & \textbf{V} & \textbf{VI} & \textbf{VII} & \textbf{VIII} & \textbf{\textit{MET.}} \\
            \midrule
            \rowcolor{cyan!07}
            LMPM~\cite{MeViS} & 25.8 & 31.2 & 28.7 & 20.0 & 22.7 & 21.3 & 20.9 & 30.0 & 31.4 & -  \\
            EEMC~\cite{refavs} & 29.6 & 34.4 & 32.6 & 19.6 & 26.0 & 28.0 & 24.7 & 35.6 & 36.0 & - \\
            \rowcolor{cyan!07}
            MUTR~\cite{mutr} & 32.3 & 35.4 & 33.3 & 28.4 & 29.8 & 26.5 & 22.8 & 41.6 & 40.5 & - \\
            LISA-7B~\cite{lisa} & 33.6 & 33.3 & 31.2 & 29.2 & 32.7 & 28.6 & 27.3 & 43.4 & 43.1 & 11.6 \\
            \rowcolor{cyan!07}
            LISA-13B~\cite{lisa} & 36.1 & 36.4 & 32.1 & 30.4 & 35.7 & 31.6 & 30.2 & 46.7 & 45.7 & 16.5 \\
            OISA-1B~\cite{OmniAVS} & \textbf{41.1} & \textbf{40.1} & \textbf{38.5} & \textbf{34.9} & \textbf{38.5} & \textbf{35.9} & \textbf{35.2} & \textbf{52.6} & \textbf{53.0} & \textbf{21.7} \\
            \bottomrule
        \end{tabular}
    }
    \label{tab:omniavs}
\end{table}

%% file: Tabs/3d.tex
\begin{table}[!t]
    \centering
    \footnotesize
    \caption{\textbf{Performance Comparison on 3D-RES Benchmarks.} 
     The results are evaluated on ScanRefer~\cite{scanrefer} dataset with Acc@K and mIoU metric. $^*$ denotes weakly-supervised methods.
    }
    \vspace{-3mm}
    \setlength{\tabcolsep}{1.6pt}
    \resizebox{0.48\textwidth}{!}{\begin{tabular}{rccccccccc}
    \toprule
    \multirow{2}{*}{\textbf{Method}} & \multirow{2}{*}{\textbf{Venue}} & \multicolumn{2}{c}{\textbf{Unique~($\sim$19\%)}} & \multicolumn{2}{c}{\textbf{Multiple~($\sim$81\%)}} & \multicolumn{2}{c} {\textbf{Overall}}  & \multirow{2}{*}{\textbf{mIoU}}\\
    \cmidrule(lr){3-4} \cmidrule(lr){5-6} \cmidrule(lr){7-8}
    & & Acc@0.25 & Acc@0.5 & Acc@0.25 & Acc@0.5 & Acc@0.25 & Acc@0.5 \\ 
    \midrule    
    \rowcolor{cyan!07}
    TGNN~\cite{tgnn} & \pub{AAAI'21} & - & - & - & - & 37.50 & 31.40 &28.80 \\
    X-RefSeg3D~\cite{x-refseg3d} & \pub{AAAI'24} & - & - & - & - & 40.33 & 33.77& 29.94\\
    \rowcolor{cyan!07}
    3D-STMN~\cite{3d-stmn} & \pub{AAAI'24} & 89.30 & 84.00 & 46.20 & 29.20 & 54.60 & 39.80 &39.50\\
    RefMask3D~\cite{refmask3d} & \pub{MM'24} & 89.55 & 84.69 & 48.09 & 40.77 & 55.87 & 49.24 &44.86 \\
    \rowcolor{cyan!07}
    MDIN~\cite{3d-gres} & \pub{MM'24} & 91.00 & 87.20 & 50.10 & 44.90 & 58.00 & 53.10 & 48.30\\
    SegPoint~\cite{segpoint}& \pub{ECCV'24} & - & - & - & - &- &- & 41.70 \\  
    MCLN~\cite{mcln} & \pub{ECCV'24} & 89.57 & 78.22 & 53.28 & 45.88 & 58.70 & 50.70 &44.72\\
    \rowcolor{cyan!07}
    RG-SAN$^*$~\cite{Rg-san} & \pub{NeurIPS'24} & 89.20 & 84.30 & \textbf{55.00}& 35.40 & \textbf{61.70} & 44.90 &44.60\\
    UniSeg3D~\cite{UniSeg3D} & \pub{NeurIPS'24} & - & - & - & - & 41.50 & 28.00 &29.60\\
    \rowcolor{cyan!07}
    LESS$^*$~\cite{less} & \pub{NeurIPS'24} & - & - & - & - & 53.23 & 29.88 &33.74\\  

    Reason3D~\cite{reason3d} & \pub{3DV'25} & 88.40 & 84.20 & 50.50 & 31.70 & 57.90 & 41.90  &42.00\\
    \rowcolor{cyan!07}
    3D-LLaVA~\cite{3D-LLaVA} & \pub{CVPR'25} & - & - & - & - &- & - & 43.30 \\
    IPDN~\cite{ipdn} & \pub{AAAI'25} & \textbf{91.50} & \textbf{88.00} & 53.10 & \textbf{47.00} & 60.60 & \textbf{54.90}& \textbf{50.20} \\
    \bottomrule
    \end{tabular}}    
    \label{tab:3dres}
\end{table}

%% file: Tabs/grex.tex
\begin{table}[!t]
    \centering
    \caption{\textbf{Performance Comparison on GRES Benchmarks. }
    The results are evaluated on gRefCOCO~\cite{gres} dataset with the cIoU and mIoU metric. $\ddag$ denotes zero-shot method.
    }
    \vspace{-3mm}
    \setlength{\tabcolsep}{3.96pt}
    \resizebox{0.48\textwidth}{!}{\begin{tabular}{rccccccc}
    \toprule
    \multirow{2}{*}{\textbf{Method}} & \multirow{2}{*}{\textbf{Venue}} & \multicolumn{2}{c}{\textbf{Val}} & \multicolumn{2}{c}{\textbf{testA}} & \multicolumn{2}{c}{\textbf{testB}} \\
    \cmidrule(lr){3-4} \cmidrule(lr){5-6} \cmidrule(lr){7-8}
    & & \textbf{cIoU} & \textbf{gIoU} & \textbf{cIoU} & \textbf{gIoU} & \textbf{cIoU} & \textbf{gIoU} \\
    \midrule
    \rowcolor{cyan!07}
    MattNet~\cite{mattnet} & \pub{CVPR'18} & 47.51 & 48.24 & 58.66 & 59.30 & 45.33 & 46.14 \\
    CRIS~\cite{cris} & \pub{ICCV'21} & 55.34 & 56.27 & 63.82 & 63.42 & 51.04 & 51.79 \\
    \rowcolor{cyan!07}
    LTS~\cite{lts} & \pub{CVPR'21} & 52.30 & 52.70 & 61.87 & 62.64 & 49.96 & 50.42 \\
    LAVT~\cite{lavt} & \pub{CVPR'22} & 57.64 & 58.40 & 65.32 & 65.90 & 55.04 & 55.83 \\
    \rowcolor{cyan!07}
    ReLA~\cite{gres} & \pub{CVPR'23} & 62.42 & 63.60 & 69.26 & 70.03 & 59.88 & 61.02 \\
    VLT~\cite{vlt} & \pub{TPAMI'23} & 52.51 & 52.00 & 62.19 & 63.20 & 50.52 & 50.88 \\
    \rowcolor{cyan!07}
    LaSagnA$^\ddag$~\cite{lasagna} & \pub{arXiv'24} & 38.10 & 32.40 & 50.40 & 47.30 & 42.10 & 38.90 \\
    CoHD~\cite{cohd} & \pub{arXiv'24} & 65.17 & 68.42 & 71.85 & 72.67 & 62.63 & 63.60 \\
    \rowcolor{cyan!07}
    HDC~\cite{hdc} & \pub{arXiv'24} & 65.42 & 68.23 & 71.60 & 72.52 & 62.79 & 63.85 \\
    MABP~\cite{li2024bring} & \pub{arXiv'24} & 65.72 & 68.86 & 71.59 & 72.81 & 62.76 & 64.04 \\
    \rowcolor{cyan!07}
    InstAlign~\cite{nguyen2024instance} & \pub{arXiv'24} & 68.94 & 74.34 & 73.22 & 74.51 & 63.88 & 65.74 \\
    LISA-13B~\cite{lisa} & \pub{CVPR'24} & 63.96 & 65.24 & 71.00 & 69.99 & 62.29 & 62.11 \\
    \rowcolor{cyan!07}
    LQMFormer~\cite{lqmformer} & \pub{CVPR'24} & 64.98 & 70.94 & - & - & - & - \\
    GSVA-13B~\cite{gsva} & \pub{CVPR'24} & 66.38 & 70.04 & 72.79 & 73.29 & 63.20 & 65.45 \\
    \rowcolor{cyan!07}
    HieA2G ~\cite{hiea2g} & \pub{AAAI'25} & 64.20 & 68.40 & 70.40 & 72.00 & 61.00 & 62.80 \\
    PSALM-G5~(ft)~\cite{Ground-V} & \pub{CVPR'25} & 68.00 & 67.30 & 75.20 & 77.30 & \textbf{73.10}& \textbf{78.90} \\
    \rowcolor{cyan!07}
    UniRES++~\cite{UniRES++}& \pub{arXiv'25} & 69.90 & 74.40 & 74.50 & 76.00 & 66.60 & 69.80 \\
    Segment Anyword~\cite{Segment_Anyword} & \pub{ICML'25} & 67.73 & 66.08 & 73.57 & 74.63 & 67.56 & 70.90 \\
    \rowcolor{cyan!07}
    RAS  ~\cite{RAS} & \pub{arXiv'25} & \textbf{70.48} & \textbf{74.64} & \textbf{76.99} & \textbf{77.45} & 67.90 & 69.42 \\
    \bottomrule
    \end{tabular}
    }
    \label{tab:gres}
\end{table}

\begin{table}[!t]
    \centering
    \caption{\textbf{Performance Comparison on GREC Task. }
    The results are evaluated on gRefCOCO dataset~\cite{gres} with Pr@($\mathrm{F_1}$=1, IoU$\geq$0.5) and N-acc. metrics.}
    \vspace{-3mm}
    \setlength{\tabcolsep}{3.96pt}
    \resizebox{0.48\textwidth}{!}{
    \begin{tabular}{rccccccc}
    \toprule
    \multirow{2}{*}{\textbf{Method}} & \multirow{2}{*}{\textbf{Venue}} & \multicolumn{2}{c}{\textbf{Val}} & \multicolumn{2}{c}{\textbf{testA}} & \multicolumn{2}{c}{\textbf{testB}} \\
    \cmidrule(lr){3-4} \cmidrule(lr){5-6} \cmidrule(lr){7-8}
    & & \textbf{Pr} & \textbf{N-acc.} & \textbf{Pr} & \textbf{N-acc.} & \textbf{Pr} & \textbf{N-acc.} \\
    \midrule
    \rowcolor{cyan!07}
    MCN~\cite{mcn}& \pub{CVPR'20} & 28.0 & 30.6 & 32.3 & 32.0 & 26.8 & 30.3 \\
    MDETR~\cite{mdetr}& \pub{ICCV'21} & 42.7 & 36.3 & 50.0 & 34.5 & 36.5 & 31.0 \\
    \rowcolor{cyan!07}
    VLT~\cite{vlt}& \pub{TPAMI'23} & 36.6 & 35.2 & 40.2 & 34.1 & 30.2 & 32.5 \\
    UNINEXT~\cite{UNINEXT}& \pub{CVPR'23} & 58.2 & 50.6 & 46.4 & 49.3 & 42.9 & 48.2 \\
    \rowcolor{cyan!07}
    Ferret~\cite{ferret} & \pub{ICLR'24} & 54.8 & 48.9 & 49.5 & 45.2 & 43.5 & 43.8 \\
    SimVG~\cite{SimVG} & \pub{NeurIPS'24} & 62.1 & 54.7 & 64.6 & 57.2 & 54.8 & 57.2 \\
    \rowcolor{cyan!07}
    Grounding Dino~\cite{groundingdino} & \pub{ECCV'24}  & - & - & 45.7 & \textbf{79.0} & 44.8 & 76.7 \\
    NGDINO~\cite{RefDrone} & \pub{arXiv'25} & - & - & 46.1 & 83.2 & 45.6 & \textbf{78.1} \\
    \rowcolor{cyan!07}
    HieA2G~\cite{hiea2g} & \pub{AAAI'25} & \textbf{67.8} & \textbf{60.3} & \textbf{66.0} & {60.1} & \textbf{56.5} & 56.0 \\
    \bottomrule
    \end{tabular}
    }
    \label{GREC}
\end{table}

%% file: Tabs/reasonseg.tex
\begin{table}[!t]
    \centering
    \caption{\textbf{Performance Comparison on ReasonSeg Benchmarks. }
    The results are evaluated on the ReasonSeg~\cite{lisa} dataset with the mIoU and cIoU metrics. `ft' denotes fine-tuning on the ReasonSeg dataset. For MLLMs training methods, SFT denotes Supervised Fine-Tuning, RL denotes Reinforcement Learning.
    }
    \vspace{-3mm}
    \setlength{\tabcolsep}{3.96pt}
    \resizebox{0.48\textwidth}{!}{\begin{tabular}{rcccccc}
    \toprule
    \multirow{2}{*}{\textbf{Method}} & \multirow{2}{*}{\textbf{Venue}} & \multirow{2}{*}{\textbf{Training}} & \multicolumn{2}{c}{\textbf{Val}} & \multicolumn{2}{c}{\textbf{testA}}  \\
    \cmidrule(lr){4-5} \cmidrule(lr){6-7}
    & & \textbf{Method} & \textbf{mIoU} & \textbf{cIoU} & \textbf{mIoU} & \textbf{cIoU}\\
    \midrule
    \multicolumn{7}{l}{\textbf{\textit{w/o Multimodel Large Language Models (MLLMs).}}}        \\
       \rowcolor{cyan!07}
    ReLA~\cite{gres}  & \pub{CVPR'23} & - & 22.4 & 19.9 & 21.3 & 22.0 \\
    X-Decoder~\cite{x-decoder}  & \pub{CVPR'23} & - & 22.6 & 17.9 & 21.7 & 16.3 \\
       \rowcolor{cyan!07}
    OVSeg~\cite{OvSeg}  & \pub{CVPR'23} & - & 28.5 & 18.6 & 26.1 & 20.8 \\
    SEEM~\cite{seem}  & \pub{NeurIPS'24} & - & 25.5 & 21.2 & 24.3 & 18.7 \\
       \rowcolor{cyan!07}
    Grounded-SAM~\cite{grounded_sam}  & \pub{arXiv'24} & - & 26.0 & 14.5 & 21.3 & 16.4 \\
    \midrule
    \multicolumn{7}{l}{\textbf{\textit{w. Multimodel Large Language Models (MLLMs).}}}        \\
       \rowcolor{cyan!07}
    LISA-13B~(ft)~\cite{lisa} & \pub{CVPR'24} & SFT & 65.0 & \textbf{72.9} & 61.3 & \textbf{62.2} \\
    LISA++-7B~(ft)~\cite{lisa++} & \pub{arXiv'23} &SFT  & 64.2 & 68.1 & 57.0 & 59.5 \\
       \rowcolor{cyan!07}
    GSVA-7B~(ft)~\cite{gsva} & \pub{CVPR'24} & SFT& 50.5 & 56.4 & - & -  \\
    LLM-Seg-7B~\cite{llm-seg}   & \pub{CVPR'24} &SFT  & 52.3 & 47.5 & - & -\\
       \rowcolor{cyan!07}
    SAM4MLLM-8B~(ft)~\cite{sam4mllm} & \pub{ECCV'24} & SFT & 58.4 & 60.4 & - & - \\
    CoReS-13B~(ft)~\cite{CoReS} & \pub{ECCV'24} & SFT & \textbf{68.1} & - & \textbf{65.5} & - \\
       \rowcolor{cyan!07}
    SegLLM-7B~\cite{segllm}   & \pub{ICLR'25} &  SFT& 57.2 & 54.3 & 52.4 & 48.4\\
    READ-7B~(ft)~\cite{READ}   & \pub{CVPR'25} & SFT & 59.8 & 67.6 & 56.8 & 59.0 \\
       \rowcolor{cyan!07}
    POPEN-7B~\cite{POPEN}    & \pub{CVPR'25} & SFT+RL & 60.2 & 64.5 & - & - \\
    Seg-Zero-7B~\cite{Seg-Zero}  & \pub{arXiv'25} & RL & 62.6 & 62.0 & 57.5 & 52.0 \\
       \rowcolor{cyan!07}
    PixelThink-7B~\cite{PixelThink}    & \pub{arXiv'25} & RL & 63.8 & 62.7 & 60.2 & 55.8 \\
    SAM-R1-7B~\cite{SAM-R1}    & \pub{arXiv'25} & RL & 64.0 & 55.8 & 60.2 & 54.3 \\
       \rowcolor{cyan!07}
    VisionReasoner-7B~\cite{VisionReasoner}    & \pub{arXiv'25} & RL & 66.3 & - & 63.6 & - \\
    \bottomrule
    \end{tabular}
    }
    \label{tab:reasonseg}
\end{table}